\documentclass[letterpaper]{article} 
\usepackage[submission]{aaai23}  
\usepackage{times}  
\usepackage{helvet}  
\usepackage{courier}  
\usepackage[hyphens]{url}  
\usepackage{graphicx} 
\usepackage{natbib}  
\usepackage{caption} 
\usepackage{algorithm}
\usepackage{algorithmic}

\usepackage{newfloat}
\usepackage{listings}

\DeclareCaptionStyle{ruled}{labelfont=normalfont,labelsep=colon,strut=off} 

\usepackage{amsfonts}
\usepackage{subcaption}
\usepackage{multirow}
\usepackage{appendix}

\floatstyle{ruled}
\newfloat{listing}{tb}{lst}{}
\floatname{listing}{Listing}
\title{Dual Progressive Transformations for Weakly Supervised Semantic Segmentation}
\author {
    Dongjian Huo\textsuperscript{\rm 1,2},
    Yukun Su\textsuperscript{\rm 1,3},
    Qingyao Wu\textsuperscript{\rm 1,4\thanks{Corresponding Author}}
}
\affiliations {
    \textsuperscript{\rm 1} School of Software and Engineering, South China University of Technology\\
    \textsuperscript{\rm 2} Key Laboratory of Big Data and Intelligent Robot, Ministry of Education\\
    \textsuperscript{\rm 3} Nanyang Technological University\\
    \textsuperscript{\rm 4} Pazhou Lab, Guangzhou, China\\
    huodongjian0603@gmail.com
}

\usepackage{bibentry}

\begin{document}

\maketitle

\begin{abstract}
Weakly supervised semantic segmentation (WSSS), which aims to mine the object regions by merely using class-level labels, is a challenging task in computer vision. The current state-of-the-art CNN-based methods usually adopt Class-Activation-Maps (CAMs) to highlight the potential areas of the object, however, they may suffer from the part-activated issues. To this end, we try an early attempt to explore the global feature attention mechanism of vision transformer in WSSS task. However, since the transformer lacks the inductive bias as in CNN models, it can not boost the performance directly and may yield the over-activated problems. To tackle these drawbacks, we propose a Convolutional Neural Networks Refined Transformer (CRT) to mine a globally complete and locally accurate class activation maps in this paper. 
To validate the effectiveness of our proposed method, extensive experiments are conducted on PASCAL VOC 2012 and CUB-200-2011 datasets.
Experimental evaluations show that our proposed CRT achieves the new state-of-the-art performance on both the weakly supervised semantic segmentation task the weakly supervised object localization task, which outperform others by a large margin. Code is available at \url{https://github.com/huodongjian0603/crt}
\end{abstract}

\section{Introduction}

Semantic segmentation is one of several basic tasks in the field of computer vision, which attempts to classify images at the pixel level. Thanks to the rapid development of deep neural networks, semantic segmentation models have achieved milestones in recent years~\cite{chen2017deeplab,su2022unified,su2022epnet}. However, training a deep learning model often requires a large number of pixel-level annotations, which leads to huge labor and time costs. To tackle this issue, many works try to train the models with self-supervised learning~\cite{hoyer2021three,su2021self,su2021modeling} and with some lower-cost and easier-to-obtain annotation data, e.g., bounding boxes~\cite{dai2015boxsup,khoreva2017simple}, scribbles~\cite{lin2016scribblesup,vernaza2017learning}, points~\cite{bearman2016s}, or image-level labels~\cite{wang2020self}. Image-level labels are relatively easy to obtain, and many large-scale image datasets have ready-made class labels. However, the accuracy of semantic segmentation using image-level labels as supervised information is still low. For this reason, this paper focuses on  weakly supervised semantic segmentation with image-level label.

\begin{figure}[t]
	\centering
	\begin{subfigure}{0.49\linewidth}
		\centering
		\includegraphics[width=0.9\linewidth]{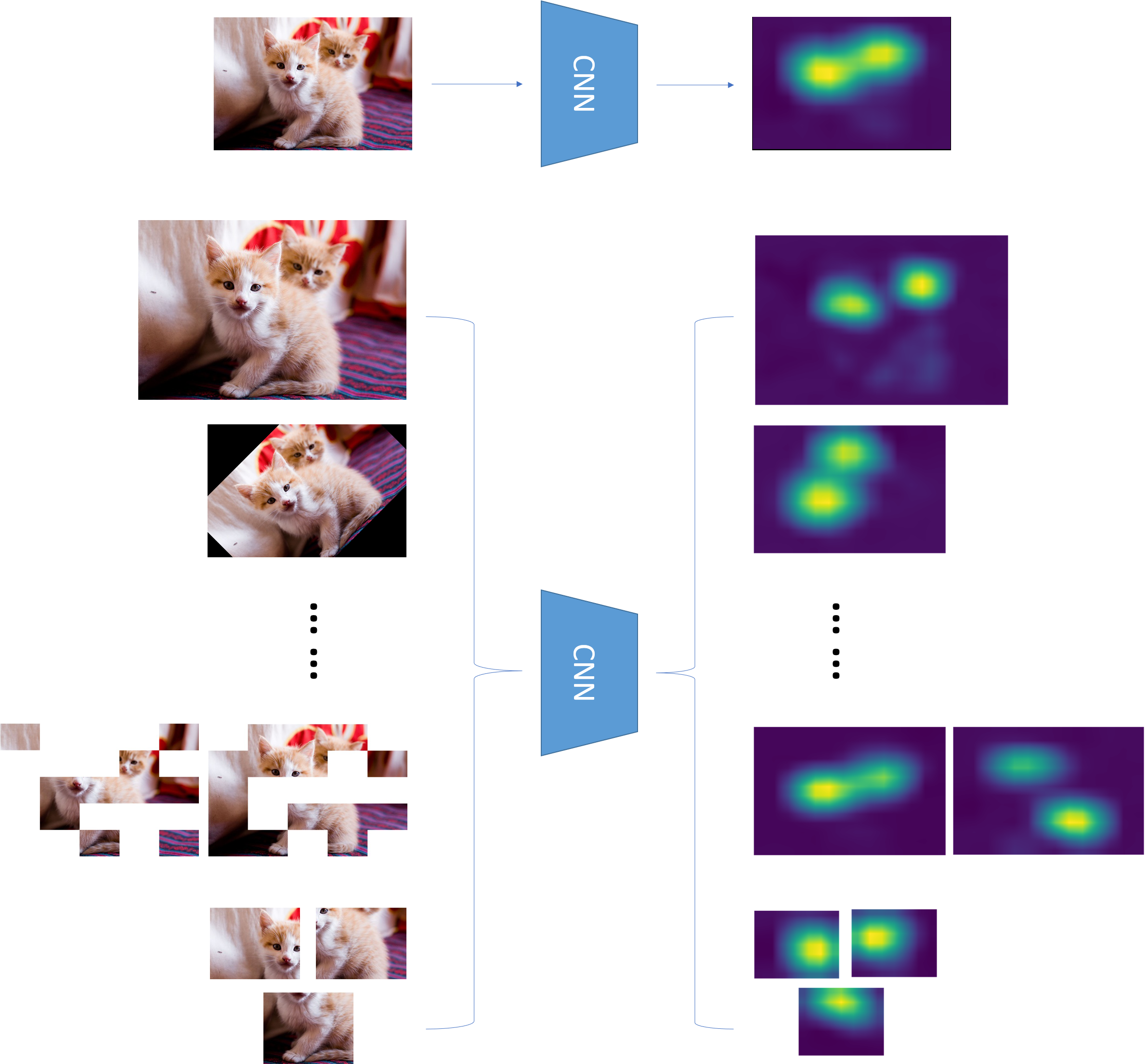}
		\caption{}
	\end{subfigure}
	\centering
	\begin{subfigure}{0.49\linewidth}
		\centering
		\includegraphics[width=0.8\linewidth]{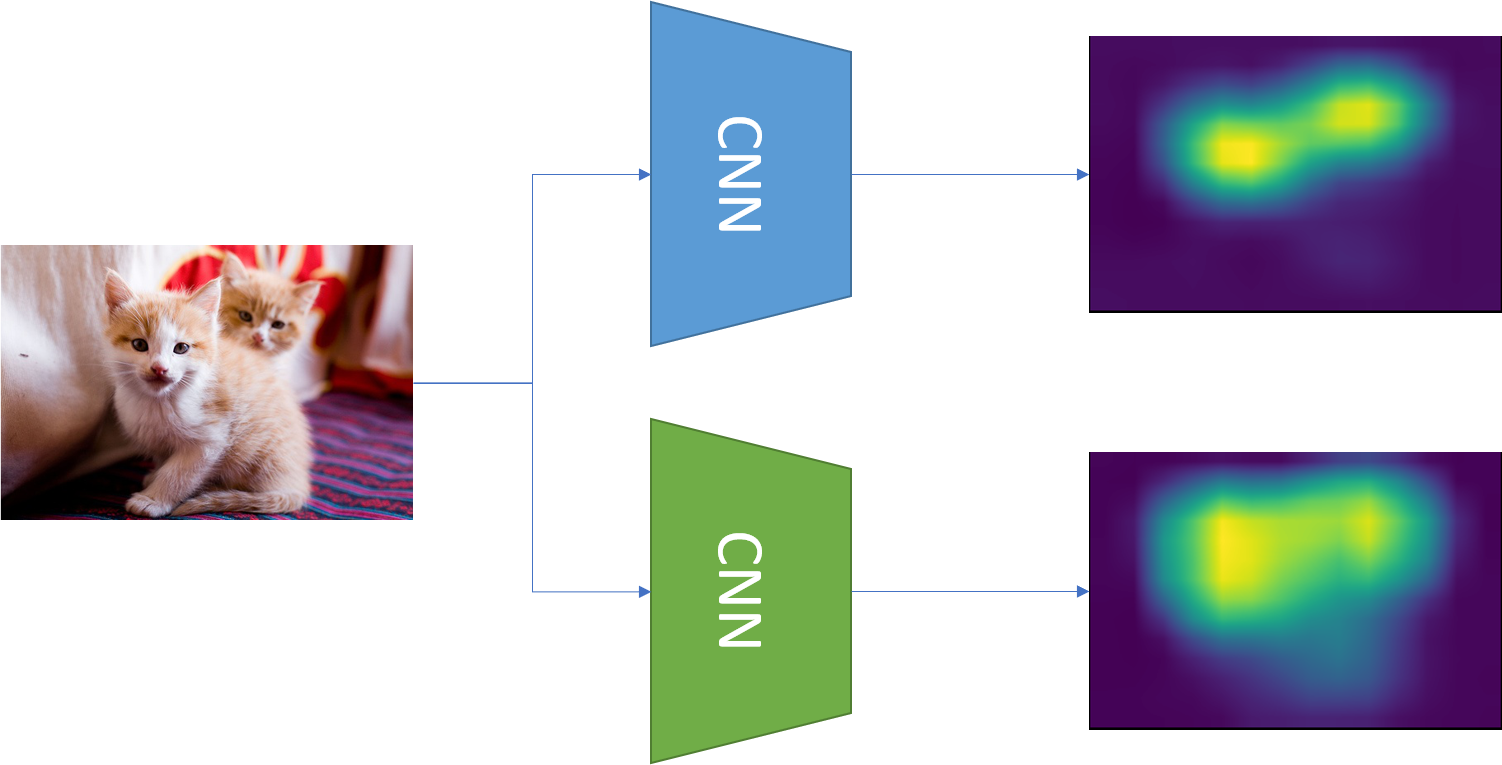}
		\caption{}
		\includegraphics[width=0.8\linewidth]{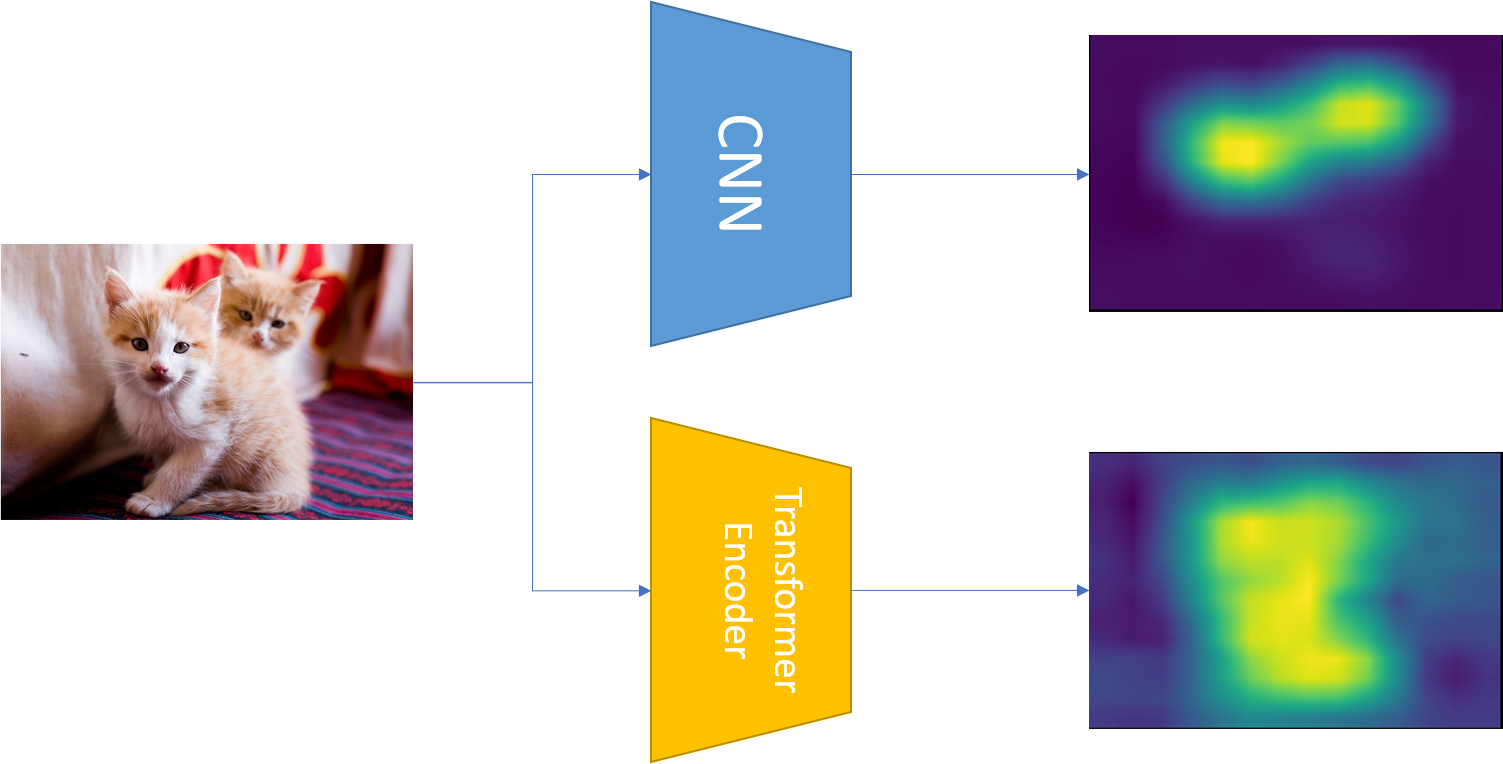}
		\caption{}
	\end{subfigure}
	\caption{Multi-branch network with the corresponding generated class activation map. (a) Both branches are CNN but using different preprocessing; (b) Both branches are CNN but using different network architecture; (c) The two branches are CNN and Transformer respectively. The CAM generated by convolution-based networks are relatively close but fail to learn  additional information from each other. For networks using different basic operations, the differences are relatively large.}
	\label{fig:intro}
\end{figure}

Most WSSS methods use the following pipeline. 1) Get the seed region through the CAM~\cite{zhou2016learning}, 2) expand the seed region to obtain pseudo-labels, 3) use pseudo-labels to train a traditional fully supervised neural network to obtain the final segmentation result. Since CAMs tend to only cover the most discriminative regions of objects and misidentify the background as the foreground, many works have been devoted to better generation of CAMs. A typical approach is to use a network with two branches to generate CAMs, and alleviate the under- and over-activation problems by minimizing the difference of CAMs~\cite{qin2022activation,wang2020self,zhang2021complementary,jiang2022l2g}. However, existing methods all use networks based on convolution operations as different branches. A shown in Figure \ref{fig:intro}, even though the two branches use slightly different CNNs, the generated class activation maps under the two views are still similar due to the same underlying operations. Using different preprocessing slightly alleviates the local activation problem, but does not fundamentally solve it. 
Other existing multi-branch methods can be regarded as a combination of the above base cases. Moreover, recent work~\cite{gao2021ts} has shown that the partial activation problem is caused by the inherent properties of CNN, where convolution operations generate local receptive fields and difficult to capture long-distance feature dependencies between pixels.

In this paper, we propose a new dual-branch network based on different basic operations to increase the difference between two views of the same image. Specifically, we choose convolution and attention mechanisms as the basic operations of the two branches. Despite the limitations of convolution, it is indeed competent for the task of CAM generation, as most existing methods have demonstrated. As for the other branch, inspired by the success of Vision Transformer (ViT)~\cite{dosovitskiy2020image} on other image tasks~\cite{gao2021ts,liu2021swin,xu2022vitpose}, we choose a ViT type network. However, in early experiments we found that ViT is not suitable for WSSS tasks for two reasons: 1) There is rich contextual information in an image, and ViT's global attention will incorrectly activate regions that do not belong to foreground objects during classification, a.k.a., the over-activation problem, 2) The size of vanilla ViT input is fixed to $224\times224$, and the size of the generated CAM is only $14\times14$, which causes a lot of information loss of the original image. Using simple interpolation or deconvolution methods, the quality of the generated CAM will be relatively low, which is not conducive to the segmentation task. In response to the problems mentioned above, we design two algorithms to enhance the ViT branch, an overlapping cutting-merging method for increasing the size of feature maps, and a residual-like correction method for mining additional semantic information.

Our method is a general approach that works well even without specific architectural changes, using networks with different underlying operations as two branches to get two distinct views of the same image, resulting in new state-of-the-art performance on the PASCAL VOC 2012 benchmark in weakly supervised semantic segmentation and the CUB-200-2011 benchmark in weakly supervised object localization(WSOL).
The contributions of this paper can be summarized as follows:

\begin{itemize}
    \item[$\bullet$]We exploit the invariance of images under different underlying operations as additional supervision for the weakly supervised semantic segmentation task, which, to the best of our knowledge, has not been well explored.
    \item[$\bullet$]We design the algorithm for the ViT branch, which alleviates the problems of over-activation and small CAM size, making it more suitable for weakly supervised semantic segmentation tasks.
    \item[$\bullet$]Our method achieves the new state-of-the-art by 71.7\% mIoU on the test set of PASCAL VOC 2012, 72.9\% Top-1 and 86.4\% Top-5 localization accuracy on the test set of CUB-200-2011.
\end{itemize}

\section{Related Work}
\subsection{Weakly supervised semantic segmentation}
Most existing WSSS methods are based on a good CAM~\cite{zhou2016learning}. The highlighted regions in the CAMs are treated as seed regions. The original seed is usually refined by learning pairwise semantic affinities~\cite{ahn2018learning,ahn2019weakly,wang2020self} to obtain pseudo segmentation masks. The pseudo segmentation mask is then used for training the final semantic segmentation model. Therefore, the class activation map largely determines the final segmentation result. However, the CAM model is trained to solve a classification task and only the most discriminative regions are activated. To this end, many works are devoted to getting a better CAM.

\hspace*{\fill} \\
\noindent {\bf Single-branch WSSS method}. Some approaches~\cite{li2018tell,kumar2017hide,wei2017object} force the model to discover other regions by suppressing the most discriminative regions by erasing or hiding. However, non-object regions tend to be incorrectly activated in the iterative update of the CAM. Another line of works~\cite{chang2020mixup,su2021context} use data augmentation strategy to calibrate the uncertainty or decouple contextual information, yet out-of-distribution data is introduced and additional training time is spent. 

\hspace*{\fill} \\
\noindent {\bf Multi-branch WSSS method}. 
Some works feed the same image with different preprocessing into the same network to alleviate the under-activation problem. SEAM~\cite{wang2020self} used two branches with shared weights to mine the invariance of images before and after affine transformation. 
Another series of work used asymmetric branches to learn more varied CAM maps. L2G~\cite{jiang2022l2g} used global and local networks, so that the global network could learn rich knowledge of local details from the local network online, leading to more complete object attention. 
AMR~\cite{qin2022activation}  utilize the spotlight branch and compensation branch to obtain weighted CAMs that provide calibration supervision and task-specific concepts. Above works alleviate the problems caused by convolution operations to a certain extent from the perspective of different image preprocessing or branch architectures. However, they do not essentially circumvent the limitations of convolution operations. Therefore, we propose a new way to solve the WSSS problem from the perspective of basic operations.
\subsection{Transformer}
\begin{figure*}[t]
\centering
\includegraphics[width=0.7\textwidth]{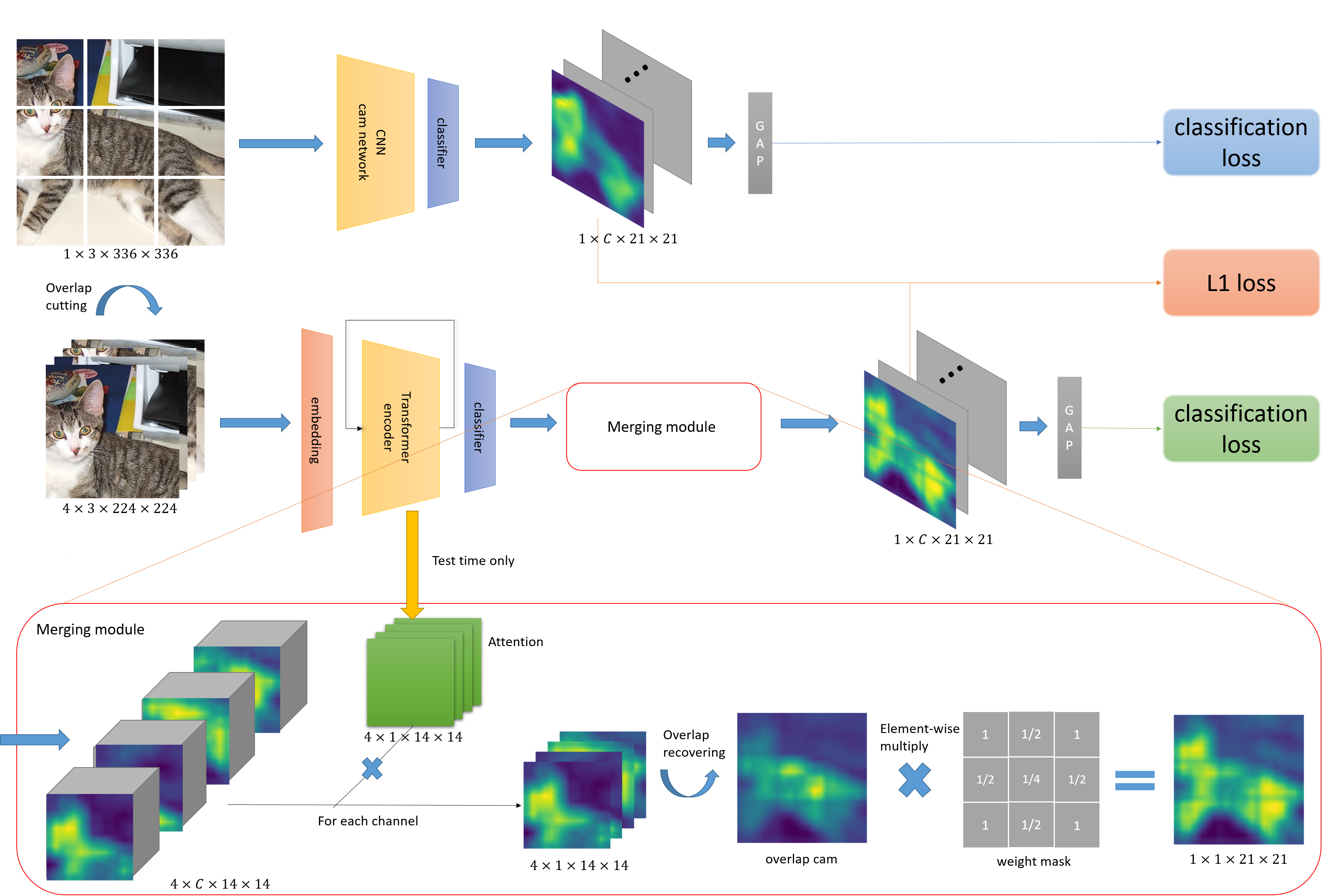} 
\caption{Overview of CRT. Using the cutting and merging module, the Transformer branch gets a higher resolution and higher quality CAM. During training, the CNN and Transformer branches co-evolve under the constraints of CAM's L1 loss. During testing, the CNN branch is discarded, only the Transformer branch is used for inference, and attention information is added in the forward process to obtain the revised CAM.}
\label{fig:overview}
\end{figure*}
Vision Transformer (ViT)~\cite{dosovitskiy2020image} proposed at the end of 2020 successfully applied the Transformer architecture to the field of image recognition for the first time, shaking the dominance of traditional CNN in computer vision. After that, a large number of Transformer-based work has emerged, achieving promising results in various vision tasks\cite{li2022exploring,liu2021swin,zheng2021rethinking,xu2022vitpose}.

DINO~\cite{caron2021emerging} found that “self-supervised ViT features contain explicit information about the semantic segmentation of an image”. However, their attention maps are class-agnostic. TS-CAM~\cite{gao2021ts} further explored the application of DeiT-S's attention map to WSOL. They re-allocate category information to patch tokens and reshape them into a semantic-aware map, and then couple semantic-agnostic attention maps and semantic-aware map to achieve semantic-aware localization. However, as we mentioned earlier, TS-CAM is insufficient for WSSS tasks. As a result, we made adaptive adjustments to it. 

\section{Methodology}

\subsection{Motivation}
Multi branch network has been widely used in weak supervised semantic segmentation, but it seems that no one has tried to use different basic operations as the basic operations of two branches to capture the invariance of CAM. Therefore, we propose a dual branch network architecture based on different operations. Specifically, we chose CNN and Transformer based network as the two branches of the overall network, as shown in Figure \ref{fig:overview}.
However, vanilla transformer is not competent for WSSS tasks, so we make task specific enhancements to the transformer branch, including an overlapping cutting-merging method for increasing the size of feature maps, and a residual-like correction method for mining additional semantic information.

\subsection{Overall architecture}
The dual branch network with different operations proposed in this paper is a pluggable method. Two branches can use any type of network architecture, and the only requirement is to use different basic operations. Following the convention and weighing the advantages and disadvantages, this paper uses ResNet152 and DeiT-S as the backbone networks of the two branches respectively.

\hspace*{\fill} \\
\noindent {\bf ResNet branch}. the ResNet used here is an improved version of the original version by IRNet~\cite{ahn2019weakly}. In order to make it more suitable for CAM generation tasks, the step size of each layer is adjusted to [2,2,2,1] by IRNet, and finally the downsampling rate is 16. We remove the global average pool layer behind the last layer of convolution layer, and connect the feature map output from the last layer of convolution layer directly with the classification head to obtain the CAM. Refer to ~\cite{ahn2019weakly,he2016deep} for other details.

\hspace*{\fill} \\ 
\noindent {\bf DeiT-S branch}. The DeiT-S used here follows ~\cite{gao2021ts,caron2021emerging}. Given a input picture, the picture is first divided into $N\times N$ patches, and then these patches are flattened and linearly projected to the embedded layer to obtain the patch tokens, $T_p\in \mathbb{R}^{N^2\times D}$, where $D$ is the embedding dimension. After  patch tokens $T_p$ are concatenated to a class token for classification, $T_{cls}\in \mathbb{R}^{1\times D}$, add a learnable position embedding to get the input tokens, $T_{in}\in \mathbb{R}^{(N^2+1)\times D}$. Then $T_{in}$ are fed into the Transformer encoder consisting of $L$ layer to obtain the output $T_{out}\in \mathbb{R}^{(N^2+1)\times D}$. Each encoding layer includes a Multi-Head Attention (MHA) module and a Multi-layer Perceptron (MLP).

Discard the token implying category information in the encoder output $T_{out}$, and reshape the remaining into token feature maps, $T_{p\_out}\in \mathbb{R}^{D \times N \times N}$. $T_{p\_out}$ enter the following classifier to get the semantic-aware CAM, $T_{cam\_out}\in \mathbb{R}^{C \times N \times N}$, where $C$ represents the number of categories of foreground objects. After Global Average Pooling (GAP) layer, $T_{cam\_out}$ is transformed into the scores of $C$ categories, and then the classification loss can be calculated.

In the phase of CAM generating, attention information is used to help mine more complete active regions. It is worth mentioning that the CAM during training is different from that at this stage. The CAM here can be regarded as the training-time CAM with refinement. From the attention module of each layer in the Transformer's encoder, we get an attention map between tokens $A_{t2t}^l\in \mathbb{R}^{(N^2+1)\times (N^2+1)}$, where $l$ is the number of layers, and $l\in \{1,2,...,L\}$. Average the inter-token attention maps of each layer to obtain the global attention map, $A_{t2t}=\frac{1}{L} \sum\nolimits_{L} A_{t2t}^l$. Further, we can get the attention map between categories and patches, $A_{c2p}\in \mathbb{R}^{(1\times N^2)}$, where $A_{c2p}=A_{t2t}[0,1:]$, and then change its shape to $A_{c2p}\in \mathbb{R}^{(N\times N)}$. Combine the semantic-agnostic attention map and the semantic-aware CAM to obtain the refined CAM: \begin{equation}\label{eq1}
T_{cam\_refine}^c=T_{cam\_out}^c\otimes A_{c2p}
\end{equation}
where $\otimes$ is the Hadamard product, $c$ is a certain category, $c\in \{0,1,2,…,C-1\}$.

\hspace*{\fill} \\
\noindent {\bf Objective function}. Assuming that the category of the dataset is $C$, for a picture of size $H\times W\times C_0$, after the same preprocessing, two copies are sent to two different branches respectively. After the original image is downsampled by $16$ times, a feature map of size $\frac{H}{16} \times \frac{W}{16} \times C_1/C_2$ is obtained, where $C_1/C_2$ stands for the number of channels of the feature maps corresponding to the two branches. After the feature maps enters the classifiers, two CAM of size $\frac{H}{16} \times \frac{W}{16} \times C$ are obtained. First calculate the $L1$ loss of the two CAMs, and then perform a global average pooling operation on each CAM to calculate a multi-label classification loss with the ground-truth label. Finally, the total loss is given by:
\begin{equation}\label{eq2}
L_{total}=L_{cls1}+L_{cls2}+\lambda L_{l1-cam}
\end{equation}
where $\lambda$ is a hyperparameter that controls the similarity of the two CAMs, which we discuss in more detail in Ablation Study.

\subsection{Enhancements for Transformer branch}
The original DeiT-S network can handle the classification task well, but it is not suitable as a CAM generation network due to problems such as over-activation and too small generated feature maps. To this end, this paper makes two algorithm designs for DeiT-S that are suitable for the field of weakly supervised semantic segmentation.

\hspace*{\fill} \\
\noindent {\bf overlapping cutting-merging}. Image segmentation task is a pixel-level task, so too small CAM will affect the quality of pseudo segmentation. The input size of the Transformer-based visual model is generally fixed at $224\times224$, and after the feature extraction network, the original image is downsampled to a $14\times14$ CAM. According to our experiments, this size of CAM image is not enough for segmentation tasks. Therefore, this paper proposes a method to increase the size of the input image and the output CAM. The specific methods are as follows: Suppose the input is $B\times3\times W\times H$ and the number of foreground categories in the dataset is $C$. (1) Traverse the entire image with a stride of 112, and cut a $224\times224$ sub-graph at each position. (2) Concatenate the obtained sub-graphs in the batch dimension. (3) Feed the processed input to the network to get the sub-CAM group. (4) Merging according to the inverse process of cutting to obtain $B\times C\times \frac{W}{16} \times \frac{H}{16}$ overlapping CAMs. (5) Do an average according to the number of repeated calculations of the overlapping part to get the real CAM map. Note that if the length or width of the original input is not a multiple of 224, then we first zero-pad the image to the bottom right to a multiple of 224.

This approach has three advantages: (1) Regardless of hardware factors, the size of the input image can theoretically be large, and the information of each pixel in the original image is not lost. During the test time, the information of the small objects in the picture is completely preserved. (2) The picture is actually divided into many different but intersecting sub-graphs, which greatly increases the diversity of the input. (3) The multi-scale strategy can be used in the stage of CAM generating, which is not possible if the cut-merge method is not used. Although this approach has various benefits, there are still problems. The number of input batches will increase in square level as the size of the image increases, resulting in more memory being occupied. To trade off accuracy and speed, we crop the input image to a size of $336\times336$ during training. In subsequent ablation experiments, we will justify this.

\begin{figure}[t]
\centering
\includegraphics[width=0.9\columnwidth]{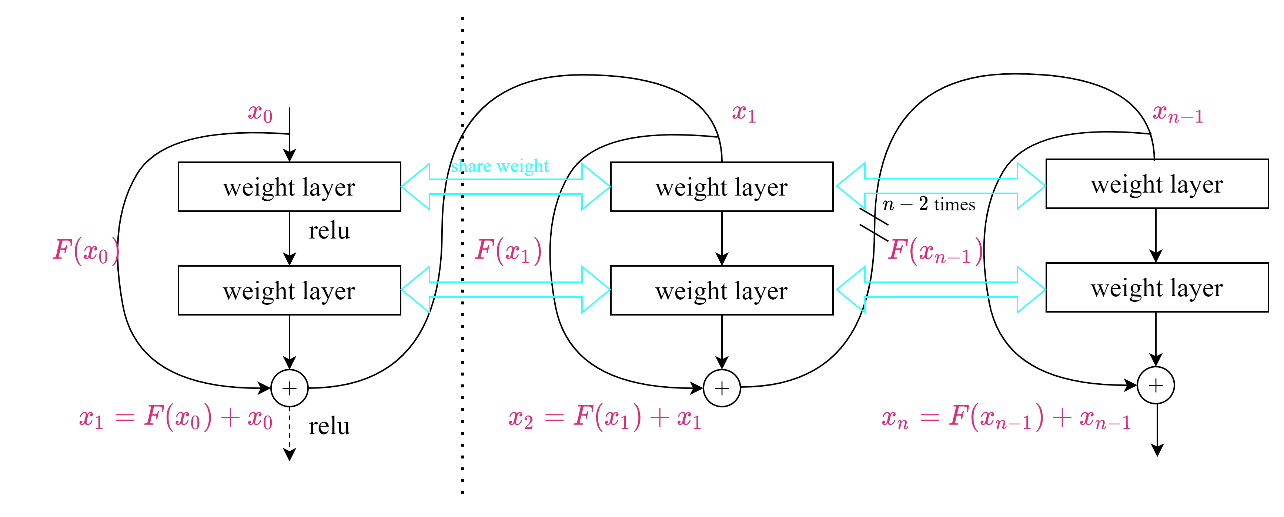} 
\caption{The difference and connection between the residual structure and ours. The left part of the dotted line is a residual module.}.
\label{fig:res}
\end{figure}

\hspace*{\fill} \\
\noindent {\bf residual-like correction}. The encoder input $T_{in}$ and output $T_{out}$ are of the same size, see previous Methodology Section. Based on this, this paper proposes a residual-like correction method, so that the final output of the encoder contains as much semantic information as possible. Assuming that the input entering the Transformer encoder for the $n$th time and its corresponding output are $T_{in}^{(n)}$ and $T_{out}^{(n)}$ respectively, we add this output matrix to the input as the next input to the Transformer encoder, i.e.,
\begin{equation}\label{eq3}
T_{in}^{(n+1)} = T_{in}^{(n)} + T_{out}^{(n)}
\end{equation}

Where, matrix addition is used here, and the total number of times n is set to 3 by default. Although the implementation of this method is quite simple, through experiments we find that this simple approach does improve the quality of the CAM generated by the DeiT branch, see Ablation Study. We try to give a reasonable explanation for this efficient method: considering the encoder as a feature extractor, then the output $T_{out}$ is equivalent to $T_{in}$ with rich semantic information. The operation of feeding $T_{out} $ as the known semantic information and the initial input $T_{in}$ into the network again is equivalent to telling the feature extractor that we already know the semantic information and hope to obtain new additional semantic information.

Although the above approach is very similar to the residual structure, and even our code is based on residuals, it is necessary to emphasize that our approach is different from residuals, and our approach can be regarded as temporal residuals structure, as shown in Figure \ref{fig:res}. The difference is that (1) From the implementation point of view, the residual structure is to splicing the input and output of the current module as the input of the next module, while our approach always inputs the same encoder. (2) From the perspective of purpose, the purpose of the residual is to transform the training objective into learning an equivalent residual mapping and ours is to learn an additional semantic extractor, hoping that the network can pay attention to places it has not paid attention to before.

\section{Experiments}

\begin{table}
\centering
\small
\begin{tabular}{c|cccc}
\hline
\makebox[0.1\textwidth][c]{$\lambda$} & 0 & 0.01 & 0.1 & 0.2 \\
\hline
\makebox[0.1\textwidth][c]{Seed} & 40.8 & 47.7 & \textbf{51.5} & 50.6 \\
\makebox[0.1\textwidth][c]{Mask} & 49.0 & 63.0 & \textbf{67.0} & 66.2 \\
\hline
\end{tabular}
\caption{The mIoU (\%) performance comparison of different values of $\lambda$ on the PASCAL VOC training set.}
\label{tab:lambda}
\end{table}

\subsection{Experimental Setup}
{\bf Datasets.} We verify the effectiveness of the proposed method on the PASCAL VOC 2012 dataset and conduct extended experiments on CUB-200-2011 dataset. The PASCAL VOC 2012 dataset, which contains 20 classes of foreground objects and 1 class of background, consists of three subsets, namely training set, validation set and test set, with 1464, 1449 and 1456 images respectively. Following the convention of previous work~\cite{ahn2019weakly,wang2020self}, we use the additional 10582 images provided in ~\cite{hariharan2011semantic} as the augmented training set. CUB-200-2011 is often used as a benchmark for WSOL tasks. The dataset consists of 200 categories of birds, divided into training and test sets, containing 5994 and 5794 images respectively.

\noindent {\bf Evaluation metric.} For WSSS task, mean intersection-over
union (mIoU) is used as a metric. And for the PASCAL VOC 2012 test set segmentation mask quality evaluation, we first generate its pseudo segmentation mask locally and submit it to the PASCAL VOC online evaluation server. For WSOL task, Top-1/Top-5 localization accuracy and Localization accuracy with ground-truth class are used.

\noindent {\bf Implementation details.} When training the classification network, the training images are resized to $210\times420$ and then randomly cropped to $336\times336$ and the hyperparameter $\lambda$ is set to $0.1$. For semantic segmentation, we use Deeplab V2~\cite{chen2017deeplab} based on ResNet101 as the semantic segmentation model. At test time, we use multi-scale testing and CRF for post-processing, where the hyperparameters of CRF are set as suggested in ~\cite{chen2017deeplab}.

\subsection{Ablation Study}

\begin{figure}[t]
\centering
\includegraphics[width=0.9\columnwidth]{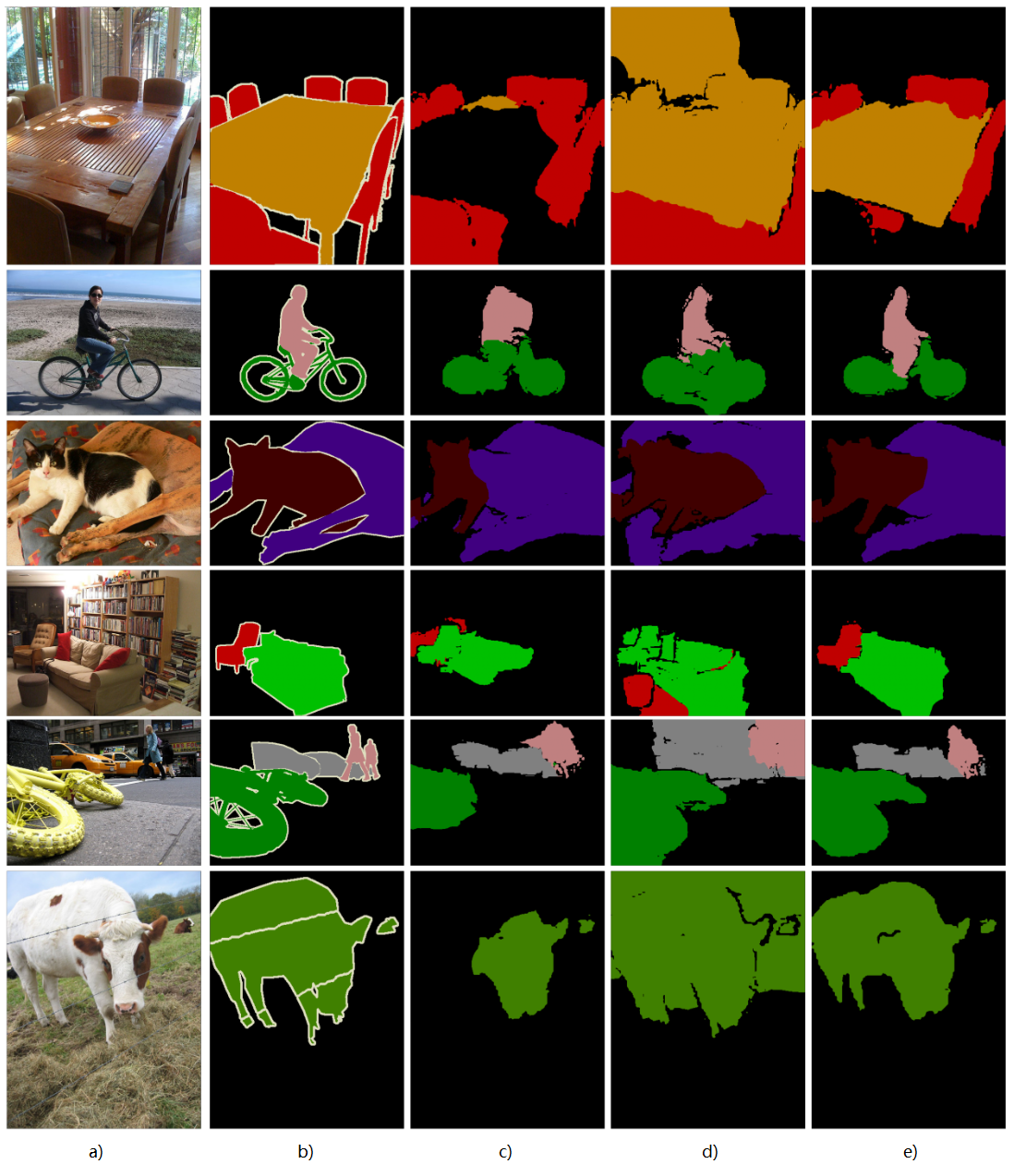} 
\caption{Visualization of pseudo-segmentation masks on the PASCAL VOC 2012 training set. a) Input image; b) Ground truth; c) IRNet; d) TS-CAM; e) CRT}
\label{fig:mask}
\end{figure}

\begin{table}
\centering
\scriptsize
\begin{tabular}{lcccc}
\hline
\makebox[0.15\columnwidth][l]{Method} & \makebox[0.1\columnwidth][c]{Input} & \makebox[0.1\columnwidth][c]{CAM} & \makebox[0.05\columnwidth][c]{Seed} & \makebox[0.05\columnwidth][c]{Mask}\\
\hline
baseline & $224\times224$ & $14\times14$ & 51.5 & 67.0\\
oC\&M & $336\times336$ & $21\times21$ & 57.6 & \textbf{69.9}\\
non-oC\&M & $448\times448$ & $28\times28$ & \textbf{57.9} & 69.4\\
bilinear interp. & $224\times224$ & $28\times28$ & 53.2 & 66.9\\
deconvolution & $224\times224$ & $28\times28$ & 29.2 & -\\
\hline
\end{tabular}
\caption{Comparison of mIoU (\%) performance of different augmented feature map methods on the PASCAL VOC training set. For the ResNet branch, the input size is 16 times the size of the feature map of the DeiT-S branch. oC\&M stands for overlapping cutting-merging}
\label{tab:augmented}
\end{table}

\begin{figure}[t]
\centering
\includegraphics[width=0.8\columnwidth]{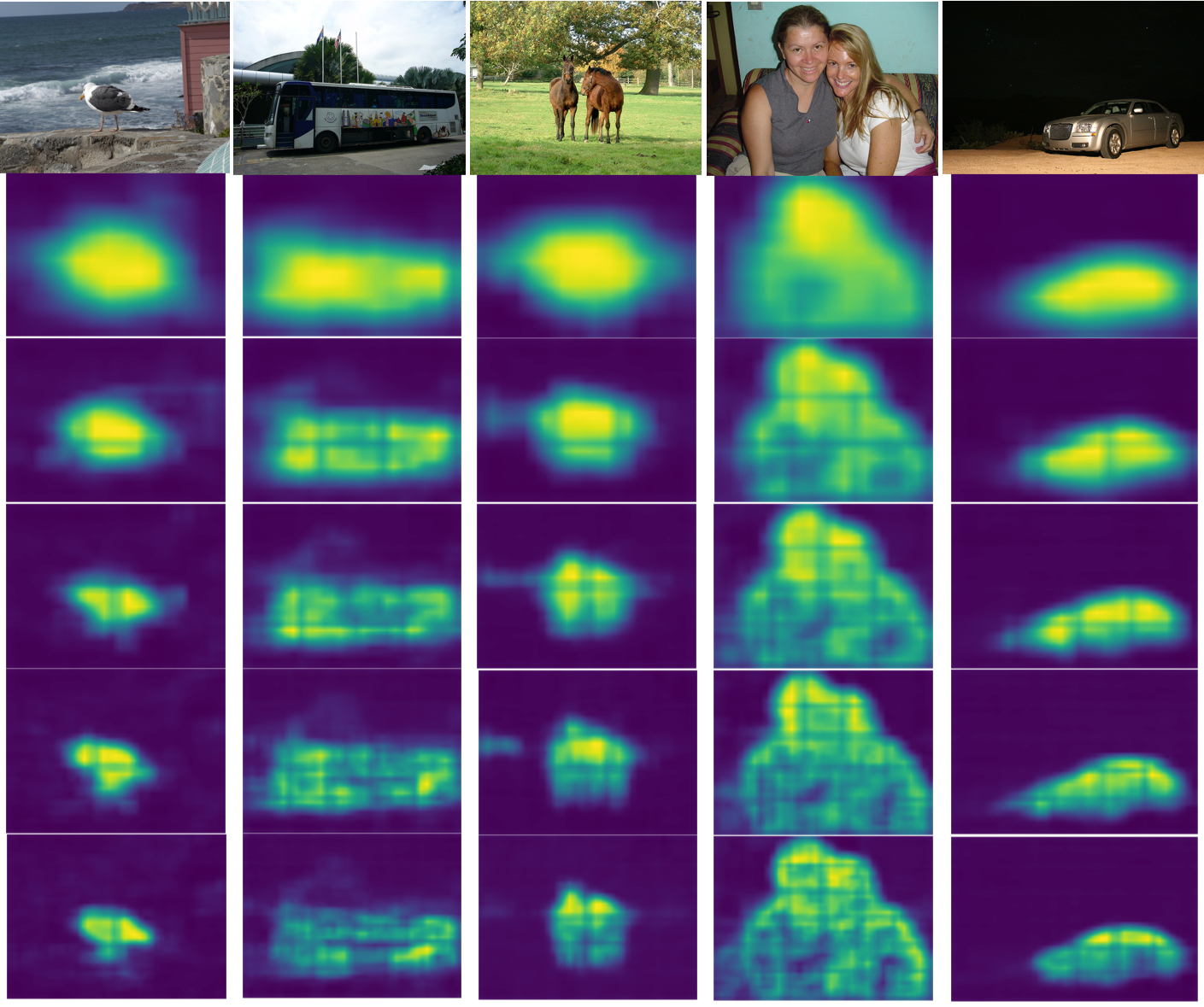} 
\caption{The effect of different input sizes on CAM quality. The input images are all taken from the PASCAL VOC 2012 training set. Among them, from top to bottom are the input pictures, and the CAMs of input size of $224\times224$, $336\times336$, $448\times448$, $560\times560$, $672\times672$.}
\label{fig:cam}
\end{figure}

{\bf The effect of hyperparameter $\lambda$.} In this subsection, we study the influence of different values of hyperparameters on the experimental results, in order to explore the hyperparameter sensitivity of the method proposed in this paper, and to find the most suitable hyperparameter settings. Note that when $\lambda=0$, it is equivalent to training a DeiT-S network as a classification network alone without using ResNet to correct it. As can be seen from Table ~\ref{tab:lambda}, using ResNet to correct the DeiT-S network, the mIoU results are better than those without using ResNet. Among them, $\lambda=0.1$ has the best effect, which is 10.7\% and 18.0\% higher than the benchmark in the seed area and the accuracy after random walk, respectively. We can infer that DeiT-S does learn useful semantic information from ResNet.

For a more intuitive analysis, we visualize the pseudo segmentations generated by IRNet, TS-CAM and CRT. As can be seen from Figure ~\ref{fig:mask}, IRNet has a problem of local activation due to the characteristics of the CNN structure itself, and only the most discriminative regions in the pseudo segmentation mask are activated. TS-CAM finds a more comprehensive area by virtue of global attention, however, the local details are far inferior to CRT, and even many areas belonging to the background are misjudged as foreground. We speculate that the huge field of view possessed by ViT enables it to make full use of contextual information to supplement judgment in classification tasks. As shown in the last line of Figure ~\ref{fig:mask}, TS-CAM regards grass and forest as foreground output, which is reasonable for the classification task of cattle, but it is not the CAM we want. As conjectured by the NFL theorem, although DeiT-S performs better on classification tasks, it is less suitable for CAM generation tasks. So our changes to the DeiT-S branch are necessary.

As we expected, the activated areas of branches using different basic operations are very different, but they have a certain consensus, that is, the CAM invariance of the same image in different perspectives. Specifically, for the same object, after the two branches learn through independent training, their respective activated regions should contain more or less the real mask of the object, and the model can improve the confidence of the intersection through mutual learning. For the background, since the basic operations of the two branches are different, the possibility of identifying the background at the same position as the foreground is almost 0. After mutual learning, the confidence of the region misjudged as the foreground decreases. In the second row of Figure ~\ref{fig:mask}, both IRNet and TS-CAM recognize a part of the background as a foreground object, but IRNet misjudges the background pixels around the person as a person, while TS-CAM misjudges the shadow of a bicycle as a bicycle , and through our proposed CRT method, these misjudged background regions will be suppressed due to the mutual learning of the two branches, resulting in a more accurate foreground segmentation mask.

\begin{table}
\centering
\small
\begin{tabular}{c|ccccc}
\hline
\makebox[0.25\columnwidth][c]{Number of loops} & 0 & 1 & 2 & 3 & 4\\
\hline
\makebox[0.25\columnwidth][c]{Mask} & 69.9 & 70.4 & 71.4 & \textbf{71.8} & 71.7 \\
\hline
\end{tabular}
\caption{Comparison of mIoU (\%) performance on PASCAL VOC training set with different cycle times. A loop count of 0 is equivalent to not using the residual-like correction method.}
\label{tab:loops}
\end{table}
\begin{figure}[t]
\centering
\includegraphics[width=0.4\columnwidth]{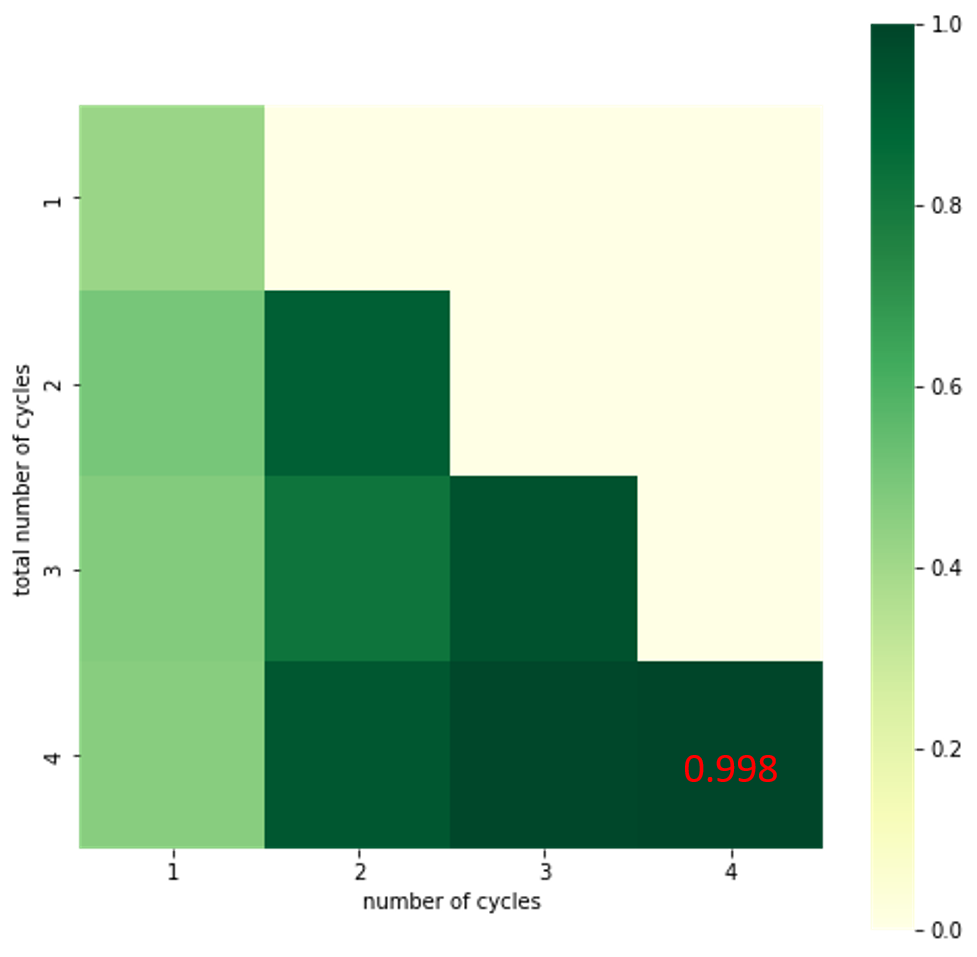} 
\caption{Cos similarity of input and output after several loops.}
\label{fig:cossim}
\end{figure}

\noindent {\bf Different ways of expanding the feature map.} In order to verify the effectiveness of the feature map enlarging method proposed in this paper, this subsection adopts a variety of different feature map enlarging methods, namely the overlapping and non-overlapping cutting-merging method, bilinear interpolation and deconvolution. As can be seen from Table ~\ref{tab:augmented}, using the cutting-merging method can indeed improve the mIoU accuracy, while using the overlapping cutting-merging method can get satisfactory results comparable to non-overlapping cutting-merging, even when the input image is smaller. Therefore, in order to reduce the overhead of the ResNet branch, we use an overlapping cutting-merging method. During testing, we use the same upsampling method as during training.

In order to further prove the rationality of our input image size as $336\times336$, we use the overlapping cutting-merging method to qualitatively compare the CAM images generated by input images of different sizes on the PASCAL VOC training set. As shown in Figure ~\ref{fig:cam}, the quality of the CAM corresponding to the input size of $224\times224$ is the worst, the outline of the foreground object is blurred, the foreground and the background cannot be clearly distinguished, and the small objects in the picture generally have the problem of over-activation. The CAM using the overlapping cutting-merging method can locate the foreground objects more accurately, and with the increase of the input size, the quality of the CAM is better.

Although the quality of the CAM with a larger size is higher, the overhead is also higher. For an input image with a size of $448\times448$, it will be divided into 9 sub-images of $224\times224$, and then sent to the network. The overhead is 9 times the baseline. For the largest size here, an input image of $672\times672$, the overhead is as high as 25 times the baseline, while the performance is only marginally better than an image with an input size of $336\times336$. Further, what is obtained here is only the seed area of the pseudo segmentation mask. The CAM only needs to roughly locate the foreground target, and then correct the seed through random walks to finally obtain the pseudo label for training the semantic segmentation model. For images with input sizes of $336\times336$ and $672\times672$, the difference after correction will be more subtle. Therefore, to balance the cost and benefit, we set the input image size to be $336\times336$.

\begin{table}
\centering
\small
\begin{tabular}{lcc}
\hline
\makebox[0.3\columnwidth][l]{Method} & \makebox[0.15\columnwidth][c]{Seed} & \makebox[0.15\columnwidth][c]{Mask} \\
\hline 
IRNet$_{CVPR}$\citeyearpar{ahn2019weakly} & 48.8 & 66.3 \\
Chang et al.$_{CVPR}$\citeyearpar{chang2020weakly} & 50.9 & 63.4 \\
SEAM$_{CVPR}$\citeyearpar{wang2020self} & 55.4 & 63.6 \\
CPN$_{ICCV}$\citeyearpar{zhang2021complementary} & 57.4 & 67.8 \\
AdvCAM$_{CVPR}$\citeyearpar{lee2021anti} & 55.6 & 69.9 \\
MCTformer$_{CVPR}$\citeyearpar{xu2022multi} & \textbf{61.7} & 69.1 \\
ReCAM$_{CVPR}$\citeyearpar{chen2022class} & 56.6 & 70.5 \\
VML-L$_{IJCV22}$\citeyearpar{ru2022weakly} & 57.3 & 71.4 \\
\hline
CRT (Ours) & 57.7 & \textbf{71.8} \\
\hline
\end{tabular}
\caption{Comparison of mIoU (\%) performance of initial seed regions generated by different weakly supervised semantic segmentation methods and their corresponding pseudo-segmentation masks on PASCAL VOC 2012 \textit{training} set.}
\label{tab:sota_train}
\end{table}

\begin{table}
\centering
\scriptsize
\begin{tabular}{lcc|cc}
\hline
\makebox[0.3\columnwidth][l]{Method} & \makebox[0.15\columnwidth][c]{Backbone} & \makebox[0.05\columnwidth][c]{Sup.} & \makebox[0.1\columnwidth][c]{\textit{val}} & \makebox[0.05\columnwidth][c]{\textit{test}} \\
\hline
Zhang et al.$_{ECCV}$\citeyearpar{zhang2020splitting} & ResNet50 & I+S	& 66.6 & 66.7 \\
Sun et al.$_{ECCV}$\citeyearpar{sun2020mining} & ResNet101 & I+S & 66.2 & 66.9 \\
Yao et al.$_{CVPR}$\citeyearpar{yao2021non} & ResNet101 & I+S & 68.3 & 68.5 \\
AuxSegNet$_{ICCV}$\citeyearpar{xu2021leveraging} & WideResNet38 & I+S & 69.0 & 68.6 \\
RIB$_{NeurIPS}$\citeyearpar{lee2021reducing} & ResNet101 & I+S & 70.2 & 70.0 \\
EDAM$_{CVPR}$\citeyearpar{wu2021embedded} & ResNet101 & I+S & 70.9	& 70.6 \\
EPS$_{CVPR}$\citeyearpar{lee2021railroad} & ResNet101 & I+S & 71.0	& \textbf{71.8} \\
L2G$_{CVPR}$\citeyearpar{jiang2022l2g} & ResNet101 & I+S & \textbf{72.1} & 71.7 \\
\hline
Zhang et al.$_{AAAI}$\citeyearpar{zhang2020reliability} & WideResNet38 & I	& 62.6 & 62.9 \\
SEAM$_{CVPR}$\citeyearpar{wang2020self} & WideResNet38	& I	& 64.5 & 65.7 \\
Chang et al.$_{CVPR}$\citeyearpar{chang2020weakly} & ResNet101 & I	& 66.1 & 65.9 \\
CONTA$_{NeurIPS}$\citeyearpar{zhang2020causal} & WideResNet38 & I & 66.1 & 66.7 \\
CDA$_{ICCV}$\citeyearpar{su2021context} & WideResNet38	& I	& 66.1 & 66.8 \\
AdvCAM$_{CVPR}$\citeyearpar{lee2021anti} & ResNet101 & I & 68.1 & 68.0 \\
ReCAM$_{CVPR}$\citeyearpar{chen2022class} & ResNet101 & I & 68.5 & 68.4 \\
CPN$_{ICCV}$\citeyearpar{zhang2021complementary} & WideResNet38 & I & 67.8	& 68.5 \\
AMN$_{CVPR}$\citeyearpar{lee2022threshold} & ResNet101 & I & 70.7 & 70.6 \\
VML-L$_{IJCV}$\citeyearpar{ru2022weakly} & ResNet101 & I & 70.6 & 70.7 \\
URN$_{AAAI}$\citeyearpar{li2022uncertainty} & ResNet101 & I & 71.2 & 71.5 \\
MCTformer$_{CVPR}$\citeyearpar{xu2022multi} & WideResNet38	& I	& \textbf{71.9} & 71.6 \\
\hline
CRT (Ours) & ResNet101 & I & 71.2 & \textbf{71.7} \\
\hline
\end{tabular}
\caption{Comparison of mIoU (\%) performance of different weakly supervised semantic segmentation methods on PASCAL VOC 2012 \textit{val} and \textit{test} sets.}
\label{tab:sota_test}
\end{table}

\begin{table}[t]
\centering
\scriptsize
\begin{tabular}{c|c|c|c|c}
\hline
\multirow{2}{*}{Method} & \multirow{2}{*}{Backbone} & \multicolumn{3}{c}{Loc. Acc} \\ \cline{3-5} & & Top-1 & Top-5 & Gt-Known \\
\hline
CAM$_{CVPR}$\citeyearpar{zhou2016learning} & GooLeNet & 41.1 & 50.7 & 55.1 \\
RCAM$_{arXiv}$\citeyearpar{zhang2020rethinking} & GooLeNet & 53.0 & - & 70.0 \\
DANet$_{ICCV}$\citeyearpar{xue2019danet} & InceptionV3 & 49.5 & 60.5 & 67.0 \\
ADL$_{CVPR}$\citeyearpar{choe2019attention} & InceptionV3 & 53.0 & - & - \\
\hline
CAM$_{CVPR}$\citeyearpar{zhou2016learning} & VGG16 & 44.2 & 52.2 & 56.0 \\
ACoL$_{CVPR}$\citeyearpar{zhang2018adversarial} & VGG16 & 45.9	& 56.5 & 59.3 \\
DANet$_{ICCV}$\citeyearpar{xue2019danet} & VGG16 & 52.5 & 62.0	& 67.7 \\
MEIL$_{CVPR}$\citeyearpar{mai2020erasing} & VGG16 & 57.5 & - & 73.8 \\
ADL$_{CVPR}$\citeyearpar{choe2019attention} & VGG16 & 52.4	& - & 75.4 \\
RCAM$_{arXiv}$\citeyearpar{zhang2020rethinking} & VGG16 & 59.0	& - & 76.3 \\
SSOL$_{AAAI}$\citeyearpar{su2022self} & InceptionV3 & -	& - & 86.3 \\
TS-CAM$_{CVPR}$\citeyearpar{gao2021ts} & Deit-S & 71.3	& 83.8 & 87.7 \\

\hline \hline
CRT(ours) & Deit-S & \textbf{72.9(+1.6)} & \textbf{86.4(+2.6)} & \textbf{90.1(+2.4)} \\
\hline
\end{tabular}
\caption{Comparison of CRT method and state-of-the-art methods on the CUB-200-2011 test set.}
\label{tab:wsol}
\end{table}

\noindent {\bf Effectiveness of residual-like correction methods.} In the default setting, the number of loops of the residual-like correction structure method is 3. In this subsection, we will first prove the effectiveness of this method, and then discuss the rationality of the number of loops. As shown in Table ~\ref{tab:loops}, the results of using the residual-like correction method are better than those of not using it, and from the overall trend, the more the number of cycles, the higher the mIoU. As shown in Figure ~\ref{fig:cossim}, with the increase of the total number of cycles, the similarity between the final output and the input is getting higher and higher. When the total number of cycles is 4, the similarity has reached 99.8\%, which is close to saturation. For the experiment with a total number of cycles of 4, the similarity between the output and the input is closer after each cycle, indicating that the network parameters are converging. From this, this paper infers that due to the strong learning ability of the DeiT-S network itself, for the same picture, after traversing all the network parameters once, and performing feature extraction again, the network can still learn new knowledge. According to the experiment, the input and output are basically the same after four cycles, and increasing the number of cycles will not have any effect, and even have the opposite effect. Therefore, in the following comparative experiments, the number of cycles is taken as 3 unless otherwise specified.

\subsection{Comparison with State-of-the-arts}
Tables ~\ref{tab:sota_train} and ~\ref{tab:sota_test} compare our method numerically with several state-of-the-art methods on the mIoU metric, and we can see that our method outperforms the state-of-the-art WSSS methods that use image categories as supervised information, even comes close to WSSS methods that use both image labels and saliency maps.
To further demonstrate the effectiveness and generality of the proposed xeno-op dual-branch architecture, we conduct additional experiments on the CUB-200-2011  dataset and compare with state-of-the-art methods for the WSOL task. We simply use a ResNet50 network to correct the same Deit-S branch as TS-CAM during training. 
As can be seen from Tables ~\ref{tab:wsol}, our method outperforms state-of-the-art WSOL methods that use image categories as supervised information.

\section{Conclusion}
This paper proposes a new dual-branch network architecture based on exclusive operations, and uses convolution and attention mechanism as the operations of the two branches, namely CRT, to verify the effectiveness of the proposed method. In addition, this paper proposes an overlapping cutting-merging method and a residual-like correction method to make the Transformer branch more suitable for weakly supervised semantic segmentation tasks. 
Experiments show that our proposed method achieves state-of-the-art on both WSSS and WSOL tasks.

\section{Acknowledgments}
This work was supported by National Natural Science Foundation of China (NSFC) 61876208 and 62272172, Key-Area Research and Development Program of Guangdong Province 2018B010108002, Pearl River S\&T Nova Program of Guangzhou 201806010081.

\bibliography{crt}

\begin{thebibliography}{61}
\providecommand{\natexlab}[1]{#1}

\bibitem[{Ahn, Cho, and Kwak(2019)}]{ahn2019weakly}
Ahn, J.; Cho, S.; and Kwak, S. 2019.
\newblock Weakly supervised learning of instance segmentation with inter-pixel
  relations.
\newblock In \emph{Proceedings of the IEEE/CVF conference on computer vision
  and pattern recognition}, 2209--2218.

\bibitem[{Ahn and Kwak(2018)}]{ahn2018learning}
Ahn, J.; and Kwak, S. 2018.
\newblock Learning pixel-level semantic affinity with image-level supervision
  for weakly supervised semantic segmentation.
\newblock In \emph{Proceedings of the IEEE conference on computer vision and
  pattern recognition}, 4981--4990.

\bibitem[{Bearman et~al.(2016)Bearman, Russakovsky, Ferrari, and
  Fei-Fei}]{bearman2016s}
Bearman, A.; Russakovsky, O.; Ferrari, V.; and Fei-Fei, L. 2016.
\newblock What’s the point: Semantic segmentation with point supervision.
\newblock In \emph{European conference on computer vision}, 549--565. Springer.

\bibitem[{Caron et~al.(2021)Caron, Touvron, Misra, J{\'e}gou, Mairal,
  Bojanowski, and Joulin}]{caron2021emerging}
Caron, M.; Touvron, H.; Misra, I.; J{\'e}gou, H.; Mairal, J.; Bojanowski, P.;
  and Joulin, A. 2021.
\newblock Emerging properties in self-supervised vision transformers.
\newblock In \emph{Proceedings of the IEEE/CVF International Conference on
  Computer Vision}, 9650--9660.

\bibitem[{Chang et~al.(2020{\natexlab{a}})Chang, Wang, Hung, Piramuthu, Tsai,
  and Yang}]{chang2020mixup}
Chang, Y.-T.; Wang, Q.; Hung, W.-C.; Piramuthu, R.; Tsai, Y.-H.; and Yang,
  M.-H. 2020{\natexlab{a}}.
\newblock Mixup-cam: Weakly-supervised semantic segmentation via uncertainty
  regularization.
\newblock \emph{arXiv preprint arXiv:2008.01201}.

\bibitem[{Chang et~al.(2020{\natexlab{b}})Chang, Wang, Hung, Piramuthu, Tsai,
  and Yang}]{chang2020weakly}
Chang, Y.-T.; Wang, Q.; Hung, W.-C.; Piramuthu, R.; Tsai, Y.-H.; and Yang,
  M.-H. 2020{\natexlab{b}}.
\newblock Weakly-supervised semantic segmentation via sub-category exploration.
\newblock In \emph{Proceedings of the IEEE/CVF Conference on Computer Vision
  and Pattern Recognition}, 8991--9000.

\bibitem[{Chen et~al.(2017)Chen, Papandreou, Kokkinos, Murphy, and
  Yuille}]{chen2017deeplab}
Chen, L.-C.; Papandreou, G.; Kokkinos, I.; Murphy, K.; and Yuille, A.~L. 2017.
\newblock Deeplab: Semantic image segmentation with deep convolutional nets,
  atrous convolution, and fully connected crfs.
\newblock \emph{IEEE transactions on pattern analysis and machine
  intelligence}, 40(4): 834--848.

\bibitem[{Chen and He(2021)}]{chen2021exploring}
Chen, X.; and He, K. 2021.
\newblock Exploring simple siamese representation learning.
\newblock In \emph{Proceedings of the IEEE/CVF Conference on Computer Vision
  and Pattern Recognition}, 15750--15758.

\bibitem[{Chen et~al.(2022)Chen, Wang, Wu, Hua, Zhang, and Sun}]{chen2022class}
Chen, Z.; Wang, T.; Wu, X.; Hua, X.-S.; Zhang, H.; and Sun, Q. 2022.
\newblock Class Re-Activation Maps for Weakly-Supervised Semantic Segmentation.
\newblock In \emph{Proceedings of the IEEE/CVF Conference on Computer Vision
  and Pattern Recognition}, 969--978.

\bibitem[{Choe and Shim(2019)}]{choe2019attention}
Choe, J.; and Shim, H. 2019.
\newblock Attention-based dropout layer for weakly supervised object
  localization.
\newblock In \emph{Proceedings of the IEEE/CVF Conference on Computer Vision
  and Pattern Recognition}, 2219--2228.

\bibitem[{Dai, He, and Sun(2015)}]{dai2015boxsup}
Dai, J.; He, K.; and Sun, J. 2015.
\newblock Boxsup: Exploiting bounding boxes to supervise convolutional networks
  for semantic segmentation.
\newblock In \emph{Proceedings of the IEEE international conference on computer
  vision}, 1635--1643.

\bibitem[{Dosovitskiy et~al.(2020)Dosovitskiy, Beyer, Kolesnikov, Weissenborn,
  Zhai, Unterthiner, Dehghani, Minderer, Heigold, Gelly
  et~al.}]{dosovitskiy2020image}
Dosovitskiy, A.; Beyer, L.; Kolesnikov, A.; Weissenborn, D.; Zhai, X.;
  Unterthiner, T.; Dehghani, M.; Minderer, M.; Heigold, G.; Gelly, S.; et~al.
  2020.
\newblock An image is worth 16x16 words: Transformers for image recognition at
  scale.
\newblock \emph{arXiv preprint arXiv:2010.11929}.

\bibitem[{Gao et~al.(2021)Gao, Wan, Pan, Peng, Tian, Han, Zhou, and
  Ye}]{gao2021ts}
Gao, W.; Wan, F.; Pan, X.; Peng, Z.; Tian, Q.; Han, Z.; Zhou, B.; and Ye, Q.
  2021.
\newblock Ts-cam: Token semantic coupled attention map for weakly supervised
  object localization.
\newblock In \emph{Proceedings of the IEEE/CVF International Conference on
  Computer Vision}, 2886--2895.

\bibitem[{Grill et~al.(2020)Grill, Strub, Altch{\'e}, Tallec, Richemond,
  Buchatskaya, Doersch, Avila~Pires, Guo, Gheshlaghi~Azar
  et~al.}]{grill2020bootstrap}
Grill, J.-B.; Strub, F.; Altch{\'e}, F.; Tallec, C.; Richemond, P.;
  Buchatskaya, E.; Doersch, C.; Avila~Pires, B.; Guo, Z.; Gheshlaghi~Azar, M.;
  et~al. 2020.
\newblock Bootstrap your own latent-a new approach to self-supervised learning.
\newblock \emph{Advances in neural information processing systems}, 33:
  21271--21284.

\bibitem[{Hariharan et~al.(2011)Hariharan, Arbel{\'a}ez, Bourdev, Maji, and
  Malik}]{hariharan2011semantic}
Hariharan, B.; Arbel{\'a}ez, P.; Bourdev, L.; Maji, S.; and Malik, J. 2011.
\newblock Semantic contours from inverse detectors.
\newblock In \emph{2011 international conference on computer vision}, 991--998.
  IEEE.

\bibitem[{He et~al.(2016)He, Zhang, Ren, and Sun}]{he2016deep}
He, K.; Zhang, X.; Ren, S.; and Sun, J. 2016.
\newblock Deep residual learning for image recognition.
\newblock In \emph{Proceedings of the IEEE conference on computer vision and
  pattern recognition}, 770--778.

\bibitem[{Hoyer et~al.(2021)Hoyer, Dai, Chen, Koring, Saha, and
  Van~Gool}]{hoyer2021three}
Hoyer, L.; Dai, D.; Chen, Y.; Koring, A.; Saha, S.; and Van~Gool, L. 2021.
\newblock Three ways to improve semantic segmentation with self-supervised
  depth estimation.
\newblock In \emph{Proceedings of the IEEE/CVF Conference on Computer Vision
  and Pattern Recognition}, 11130--11140.

\bibitem[{Jain et~al.(2021)Jain, Singh, Orlov, Huang, Li, Walton, and
  Shi}]{jain2021semask}
Jain, J.; Singh, A.; Orlov, N.; Huang, Z.; Li, J.; Walton, S.; and Shi, H.
  2021.
\newblock Semask: Semantically masked transformers for semantic segmentation.
\newblock \emph{arXiv preprint arXiv:2112.12782}.

\bibitem[{Jiang et~al.(2022)Jiang, Yang, Hou, and Wei}]{jiang2022l2g}
Jiang, P.-T.; Yang, Y.; Hou, Q.; and Wei, Y. 2022.
\newblock L2G: A Simple Local-to-Global Knowledge Transfer Framework for Weakly
  Supervised Semantic Segmentation.
\newblock In \emph{Proceedings of the IEEE/CVF Conference on Computer Vision
  and Pattern Recognition}, 16886--16896.

\bibitem[{Jo and Yu(2021)}]{jo2021puzzle}
Jo, S.; and Yu, I.-J. 2021.
\newblock Puzzle-cam: Improved localization via matching partial and full
  features.
\newblock In \emph{2021 IEEE International Conference on Image Processing
  (ICIP)}, 639--643. IEEE.

\bibitem[{Khoreva et~al.(2017)Khoreva, Benenson, Hosang, Hein, and
  Schiele}]{khoreva2017simple}
Khoreva, A.; Benenson, R.; Hosang, J.; Hein, M.; and Schiele, B. 2017.
\newblock Simple does it: Weakly supervised instance and semantic segmentation.
\newblock In \emph{Proceedings of the IEEE conference on computer vision and
  pattern recognition}, 876--885.

\bibitem[{Kumar~Singh and Jae~Lee(2017)}]{kumar2017hide}
Kumar~Singh, K.; and Jae~Lee, Y. 2017.
\newblock Hide-and-seek: Forcing a network to be meticulous for
  weakly-supervised object and action localization.
\newblock In \emph{Proceedings of the IEEE International Conference on Computer
  Vision}, 3524--3533.

\bibitem[{Lee et~al.(2021{\natexlab{a}})Lee, Choi, Mok, and
  Yoon}]{lee2021reducing}
Lee, J.; Choi, J.; Mok, J.; and Yoon, S. 2021{\natexlab{a}}.
\newblock Reducing information bottleneck for weakly supervised semantic
  segmentation.
\newblock \emph{Advances in Neural Information Processing Systems}, 34:
  27408--27421.

\bibitem[{Lee, Kim, and Yoon(2021)}]{lee2021anti}
Lee, J.; Kim, E.; and Yoon, S. 2021.
\newblock Anti-adversarially manipulated attributions for weakly and
  semi-supervised semantic segmentation.
\newblock In \emph{Proceedings of the IEEE/CVF Conference on Computer Vision
  and Pattern Recognition}, 4071--4080.

\bibitem[{Lee, Kim, and Shim(2022)}]{lee2022threshold}
Lee, M.; Kim, D.; and Shim, H. 2022.
\newblock Threshold Matters in WSSS: Manipulating the Activation for the Robust
  and Accurate Segmentation Model Against Thresholds.
\newblock In \emph{Proceedings of the IEEE/CVF Conference on Computer Vision
  and Pattern Recognition}, 4330--4339.

\bibitem[{Lee et~al.(2021{\natexlab{b}})Lee, Lee, Lee, and
  Shim}]{lee2021railroad}
Lee, S.; Lee, M.; Lee, J.; and Shim, H. 2021{\natexlab{b}}.
\newblock Railroad is not a train: Saliency as pseudo-pixel supervision for
  weakly supervised semantic segmentation.
\newblock In \emph{Proceedings of the IEEE/CVF conference on computer vision
  and pattern recognition}, 5495--5505.

\bibitem[{Li et~al.(2018)Li, Wu, Peng, Ernst, and Fu}]{li2018tell}
Li, K.; Wu, Z.; Peng, K.-C.; Ernst, J.; and Fu, Y. 2018.
\newblock Tell me where to look: Guided attention inference network.
\newblock In \emph{Proceedings of the IEEE Conference on Computer Vision and
  Pattern Recognition}, 9215--9223.

\bibitem[{Li et~al.(2022{\natexlab{a}})Li, Duan, Kuang, Chen, Zhang, and
  Li}]{li2022uncertainty}
Li, Y.; Duan, Y.; Kuang, Z.; Chen, Y.; Zhang, W.; and Li, X.
  2022{\natexlab{a}}.
\newblock Uncertainty estimation via response scaling for pseudo-mask noise
  mitigation in weakly-supervised semantic segmentation.
\newblock In \emph{Proceedings of the AAAI Conference on Artificial
  Intelligence}, volume~36, 1447--1455.

\bibitem[{Li et~al.(2022{\natexlab{b}})Li, Mao, Girshick, and
  He}]{li2022exploring}
Li, Y.; Mao, H.; Girshick, R.; and He, K. 2022{\natexlab{b}}.
\newblock Exploring plain vision transformer backbones for object detection.
\newblock \emph{arXiv preprint arXiv:2203.16527}.

\bibitem[{Lin et~al.(2016)Lin, Dai, Jia, He, and Sun}]{lin2016scribblesup}
Lin, D.; Dai, J.; Jia, J.; He, K.; and Sun, J. 2016.
\newblock Scribblesup: Scribble-supervised convolutional networks for semantic
  segmentation.
\newblock In \emph{Proceedings of the IEEE conference on computer vision and
  pattern recognition}, 3159--3167.

\bibitem[{Liu et~al.(2021)Liu, Lin, Cao, Hu, Wei, Zhang, Lin, and
  Guo}]{liu2021swin}
Liu, Z.; Lin, Y.; Cao, Y.; Hu, H.; Wei, Y.; Zhang, Z.; Lin, S.; and Guo, B.
  2021.
\newblock Swin transformer: Hierarchical vision transformer using shifted
  windows.
\newblock In \emph{Proceedings of the IEEE/CVF International Conference on
  Computer Vision}, 10012--10022.

\bibitem[{Mai, Yang, and Luo(2020)}]{mai2020erasing}
Mai, J.; Yang, M.; and Luo, W. 2020.
\newblock Erasing integrated learning: A simple yet effective approach for
  weakly supervised object localization.
\newblock In \emph{Proceedings of the IEEE/CVF conference on computer vision
  and pattern recognition}, 8766--8775.

\bibitem[{Papandreou et~al.(2015)Papandreou, Chen, Murphy, and
  Yuille}]{papandreou2015weakly}
Papandreou, G.; Chen, L.-C.; Murphy, K.~P.; and Yuille, A.~L. 2015.
\newblock Weakly-and semi-supervised learning of a deep convolutional network
  for semantic image segmentation.
\newblock In \emph{Proceedings of the IEEE international conference on computer
  vision}, 1742--1750.

\bibitem[{Qin et~al.(2022)Qin, Wu, Xiao, Li, and Wang}]{qin2022activation}
Qin, J.; Wu, J.; Xiao, X.; Li, L.; and Wang, X. 2022.
\newblock Activation Modulation and Recalibration Scheme for Weakly Supervised
  Semantic Segmentation.
\newblock In \emph{Proceedings of the AAAI Conference on Artificial
  Intelligence}, volume~36, 2117--2125.

\bibitem[{Ru et~al.(2022)Ru, Du, Zhan, and Wu}]{ru2022weakly}
Ru, L.; Du, B.; Zhan, Y.; and Wu, C. 2022.
\newblock Weakly-Supervised Semantic Segmentation with Visual Words Learning
  and Hybrid Pooling.
\newblock \emph{International Journal of Computer Vision}, 130(4): 1127--1144.

\bibitem[{Su et~al.(2022{\natexlab{a}})Su, Su, Zhang, Yang, Huang, and
  Wu}]{su2022epnet}
Su, J.; Su, Y.; Zhang, Y.; Yang, W.; Huang, H.; and Wu, Q. 2022{\natexlab{a}}.
\newblock EpNet: Power lines foreign object detection with Edge Proposal
  Network and data composition.
\newblock \emph{Knowledge-Based Systems}, 249: 108857.

\bibitem[{Su et~al.(2022{\natexlab{b}})Su, Deng, Sun, Lin, and
  Wu}]{su2022unified}
Su, Y.; Deng, J.; Sun, R.; Lin, G.; and Wu, Q. 2022{\natexlab{b}}.
\newblock A Unified Transformer Framework for Group-based Segmentation:
  Co-Segmentation, Co-Saliency Detection and Video Salient Object Detection.
\newblock \emph{arXiv preprint arXiv:2203.04708}.

\bibitem[{Su et~al.(2022{\natexlab{c}})Su, Lin, Hao, Cao, Wang, and
  Wu}]{su2022self}
Su, Y.; Lin, G.; Hao, Y.; Cao, Y.; Wang, W.; and Wu, Q. 2022{\natexlab{c}}.
\newblock Self-supervised object localization with joint graph partition.
\newblock In \emph{Proceedings of the AAAI Conference on Artificial
  Intelligence}, volume~36, 2289--2297.

\bibitem[{Su et~al.(2021{\natexlab{a}})Su, Lin, Sun, Hao, and
  Wu}]{su2021modeling}
Su, Y.; Lin, G.; Sun, R.; Hao, Y.; and Wu, Q. 2021{\natexlab{a}}.
\newblock Modeling the uncertainty for self-supervised 3d skeleton action
  representation learning.
\newblock In \emph{Proceedings of the 29th ACM International Conference on
  Multimedia}, 769--778.

\bibitem[{Su, Lin, and Wu(2021)}]{su2021self}
Su, Y.; Lin, G.; and Wu, Q. 2021.
\newblock Self-supervised 3d skeleton action representation learning with
  motion consistency and continuity.
\newblock In \emph{Proceedings of the IEEE/CVF international conference on
  computer vision}, 13328--13338.

\bibitem[{Su et~al.(2021{\natexlab{b}})Su, Sun, Lin, and Wu}]{su2021context}
Su, Y.; Sun, R.; Lin, G.; and Wu, Q. 2021{\natexlab{b}}.
\newblock Context decoupling augmentation for weakly supervised semantic
  segmentation.
\newblock In \emph{Proceedings of the IEEE/CVF international conference on
  computer vision}, 7004--7014.

\bibitem[{Sun et~al.(2020)Sun, Wang, Dai, and Van~Gool}]{sun2020mining}
Sun, G.; Wang, W.; Dai, J.; and Van~Gool, L. 2020.
\newblock Mining cross-image semantics for weakly supervised semantic
  segmentation.
\newblock In \emph{European conference on computer vision}, 347--365. Springer.

\bibitem[{Touvron et~al.(2021)Touvron, Cord, Douze, Massa, Sablayrolles, and
  J{\'e}gou}]{touvron2021training}
Touvron, H.; Cord, M.; Douze, M.; Massa, F.; Sablayrolles, A.; and J{\'e}gou,
  H. 2021.
\newblock Training data-efficient image transformers \& distillation through
  attention.
\newblock In \emph{International Conference on Machine Learning}, 10347--10357.
  PMLR.

\bibitem[{Vaswani et~al.(2017)Vaswani, Shazeer, Parmar, Uszkoreit, Jones,
  Gomez, Kaiser, and Polosukhin}]{vaswani2017attention}
Vaswani, A.; Shazeer, N.; Parmar, N.; Uszkoreit, J.; Jones, L.; Gomez, A.~N.;
  Kaiser, {\L}.; and Polosukhin, I. 2017.
\newblock Attention is all you need.
\newblock \emph{Advances in neural information processing systems}, 30.

\bibitem[{Vernaza and Chandraker(2017)}]{vernaza2017learning}
Vernaza, P.; and Chandraker, M. 2017.
\newblock Learning random-walk label propagation for weakly-supervised semantic
  segmentation.
\newblock In \emph{Proceedings of the IEEE conference on computer vision and
  pattern recognition}, 7158--7166.

\bibitem[{Wang et~al.(2020)Wang, Zhang, Kan, Shan, and Chen}]{wang2020self}
Wang, Y.; Zhang, J.; Kan, M.; Shan, S.; and Chen, X. 2020.
\newblock Self-supervised equivariant attention mechanism for weakly supervised
  semantic segmentation.
\newblock In \emph{Proceedings of the IEEE/CVF Conference on Computer Vision
  and Pattern Recognition}, 12275--12284.

\bibitem[{Wei et~al.(2017)Wei, Feng, Liang, Cheng, Zhao, and
  Yan}]{wei2017object}
Wei, Y.; Feng, J.; Liang, X.; Cheng, M.-M.; Zhao, Y.; and Yan, S. 2017.
\newblock Object region mining with adversarial erasing: A simple
  classification to semantic segmentation approach.
\newblock In \emph{Proceedings of the IEEE conference on computer vision and
  pattern recognition}, 1568--1576.

\bibitem[{Wu et~al.(2021)Wu, Huang, Gao, Wei, Wei, Luo, and
  Liu}]{wu2021embedded}
Wu, T.; Huang, J.; Gao, G.; Wei, X.; Wei, X.; Luo, X.; and Liu, C.~H. 2021.
\newblock Embedded discriminative attention mechanism for weakly supervised
  semantic segmentation.
\newblock In \emph{Proceedings of the IEEE/CVF Conference on Computer Vision
  and Pattern Recognition}, 16765--16774.

\bibitem[{Xu et~al.(2021)Xu, Ouyang, Bennamoun, Boussaid, Sohel, and
  Xu}]{xu2021leveraging}
Xu, L.; Ouyang, W.; Bennamoun, M.; Boussaid, F.; Sohel, F.; and Xu, D. 2021.
\newblock Leveraging auxiliary tasks with affinity learning for weakly
  supervised semantic segmentation.
\newblock In \emph{Proceedings of the IEEE/CVF International Conference on
  Computer Vision}, 6984--6993.

\bibitem[{Xu et~al.(2022{\natexlab{a}})Xu, Ouyang, Bennamoun, Boussaid, and
  Xu}]{xu2022multi}
Xu, L.; Ouyang, W.; Bennamoun, M.; Boussaid, F.; and Xu, D. 2022{\natexlab{a}}.
\newblock Multi-class Token Transformer for Weakly Supervised Semantic
  Segmentation.
\newblock In \emph{Proceedings of the IEEE/CVF Conference on Computer Vision
  and Pattern Recognition}, 4310--4319.

\bibitem[{Xu et~al.(2022{\natexlab{b}})Xu, Zhang, Zhang, and
  Tao}]{xu2022vitpose}
Xu, Y.; Zhang, J.; Zhang, Q.; and Tao, D. 2022{\natexlab{b}}.
\newblock ViTPose: Simple Vision Transformer Baselines for Human Pose
  Estimation.
\newblock \emph{arXiv preprint arXiv:2204.12484}.

\bibitem[{Xue et~al.(2019)Xue, Liu, Wan, Jiao, Ji, and Ye}]{xue2019danet}
Xue, H.; Liu, C.; Wan, F.; Jiao, J.; Ji, X.; and Ye, Q. 2019.
\newblock Danet: Divergent activation for weakly supervised object
  localization.
\newblock In \emph{Proceedings of the IEEE/CVF International Conference on
  Computer Vision}, 6589--6598.

\bibitem[{Yao et~al.(2021)Yao, Chen, Xie, Zhang, Shen, Wu, Tang, and
  Zhang}]{yao2021non}
Yao, Y.; Chen, T.; Xie, G.-S.; Zhang, C.; Shen, F.; Wu, Q.; Tang, Z.; and
  Zhang, J. 2021.
\newblock Non-salient region object mining for weakly supervised semantic
  segmentation.
\newblock In \emph{Proceedings of the IEEE/CVF Conference on Computer Vision
  and Pattern Recognition}, 2623--2632.

\bibitem[{Zhang et~al.(2020{\natexlab{a}})Zhang, Xiao, Wei, Sun, and
  Huang}]{zhang2020reliability}
Zhang, B.; Xiao, J.; Wei, Y.; Sun, M.; and Huang, K. 2020{\natexlab{a}}.
\newblock Reliability does matter: An end-to-end weakly supervised semantic
  segmentation approach.
\newblock In \emph{Proceedings of the AAAI Conference on Artificial
  Intelligence}, volume~34, 12765--12772.

\bibitem[{Zhang et~al.(2020{\natexlab{b}})Zhang, Zhang, Tang, Hua, and
  Sun}]{zhang2020causal}
Zhang, D.; Zhang, H.; Tang, J.; Hua, X.-S.; and Sun, Q. 2020{\natexlab{b}}.
\newblock Causal intervention for weakly-supervised semantic segmentation.
\newblock \emph{Advances in Neural Information Processing Systems}, 33:
  655--666.

\bibitem[{Zhang et~al.(2021)Zhang, Gu, Zhang, and Dai}]{zhang2021complementary}
Zhang, F.; Gu, C.; Zhang, C.; and Dai, Y. 2021.
\newblock Complementary patch for weakly supervised semantic segmentation.
\newblock In \emph{Proceedings of the IEEE/CVF International Conference on
  Computer Vision}, 7242--7251.

\bibitem[{Zhang et~al.(2020{\natexlab{c}})Zhang, Lin, Liu, Cai, and
  Kot}]{zhang2020splitting}
Zhang, T.; Lin, G.; Liu, W.; Cai, J.; and Kot, A. 2020{\natexlab{c}}.
\newblock Splitting vs. merging: Mining object regions with discrepancy and
  intersection loss for weakly supervised semantic segmentation.
\newblock In \emph{European Conference on Computer Vision}, 663--679. Springer.

\bibitem[{Zhang et~al.(2018)Zhang, Wei, Feng, Yang, and
  Huang}]{zhang2018adversarial}
Zhang, X.; Wei, Y.; Feng, J.; Yang, Y.; and Huang, T.~S. 2018.
\newblock Adversarial complementary learning for weakly supervised object
  localization.
\newblock In \emph{Proceedings of the IEEE conference on computer vision and
  pattern recognition}, 1325--1334.

\bibitem[{Zhang et~al.(2020{\natexlab{d}})Zhang, Wei, Yang, and
  Wu}]{zhang2020rethinking}
Zhang, X.; Wei, Y.; Yang, Y.; and Wu, F. 2020{\natexlab{d}}.
\newblock Rethinking localization map: Towards accurate object perception with
  self-enhancement maps.
\newblock \emph{arXiv preprint arXiv:2006.05220}.

\bibitem[{Zheng et~al.(2021)Zheng, Lu, Zhao, Zhu, Luo, Wang, Fu, Feng, Xiang,
  Torr et~al.}]{zheng2021rethinking}
Zheng, S.; Lu, J.; Zhao, H.; Zhu, X.; Luo, Z.; Wang, Y.; Fu, Y.; Feng, J.;
  Xiang, T.; Torr, P.~H.; et~al. 2021.
\newblock Rethinking semantic segmentation from a sequence-to-sequence
  perspective with transformers.
\newblock In \emph{Proceedings of the IEEE/CVF conference on computer vision
  and pattern recognition}, 6881--6890.

\bibitem[{Zhou et~al.(2016)Zhou, Khosla, Lapedriza, Oliva, and
  Torralba}]{zhou2016learning}
Zhou, B.; Khosla, A.; Lapedriza, A.; Oliva, A.; and Torralba, A. 2016.
\newblock Learning deep features for discriminative localization.
\newblock In \emph{Proceedings of the IEEE conference on computer vision and
  pattern recognition}, 2921--2929.

\end{thebibliography}

\end{document}


\maketitle

\section{I. Effectiveness of co-learning}
We found that the multi-branch networks used by the existing methods are limited by a single kind of basic operation. Their different branches focus on similar regions, which makes them unable to learn more information from each other. Therefore, we propose CRT network using different operations. 

To verify that different branches indeed benefit from our CRT method, we conduct experiments on the PASCAL VOC 2012 training set. In the experiments, we keep the same settings except for the hyperparameter $\lambda$, and train two multi-branch networks. After training, we performed a visual comparison in the CAM generation stage.

\begin{figure}[t]
\begin{center}
\includegraphics[width=0.8\columnwidth]{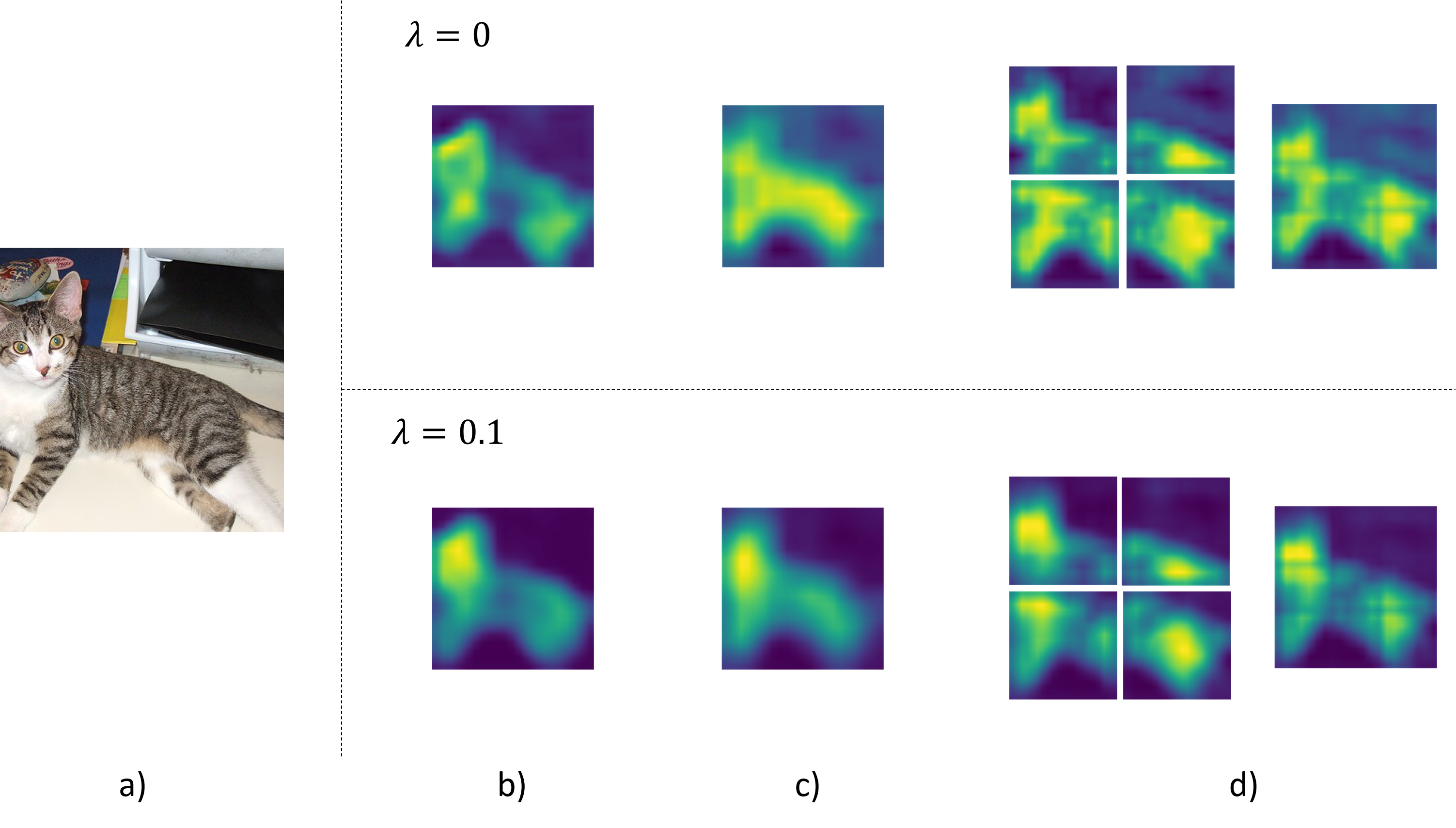}
\caption{Independent training versus mutual learning. a) Input image; b) CAM of the CNN branch; c) CAM of the Transformer branch; d) CAM of the Transformer branch, using oC\&M method.}
\label{fig:supp_mutual}
\end{center}
\end{figure}

It can be seen from Figure \ref{fig:supp_mutual} that when $\lambda=0$ (note that $\lambda=0$ is equivalent to independent training of two branches without any information exchange), both branches of the network have the problem of over-activation. This problem is especially severe on the Transformer branch. While $\lambda\neq0$, the over-activation problem is significantly suppressed.

\section{II. Additional qualitative experiments}

\begin{figure}[t]
\begin{center}
\includegraphics[width=0.8\columnwidth]{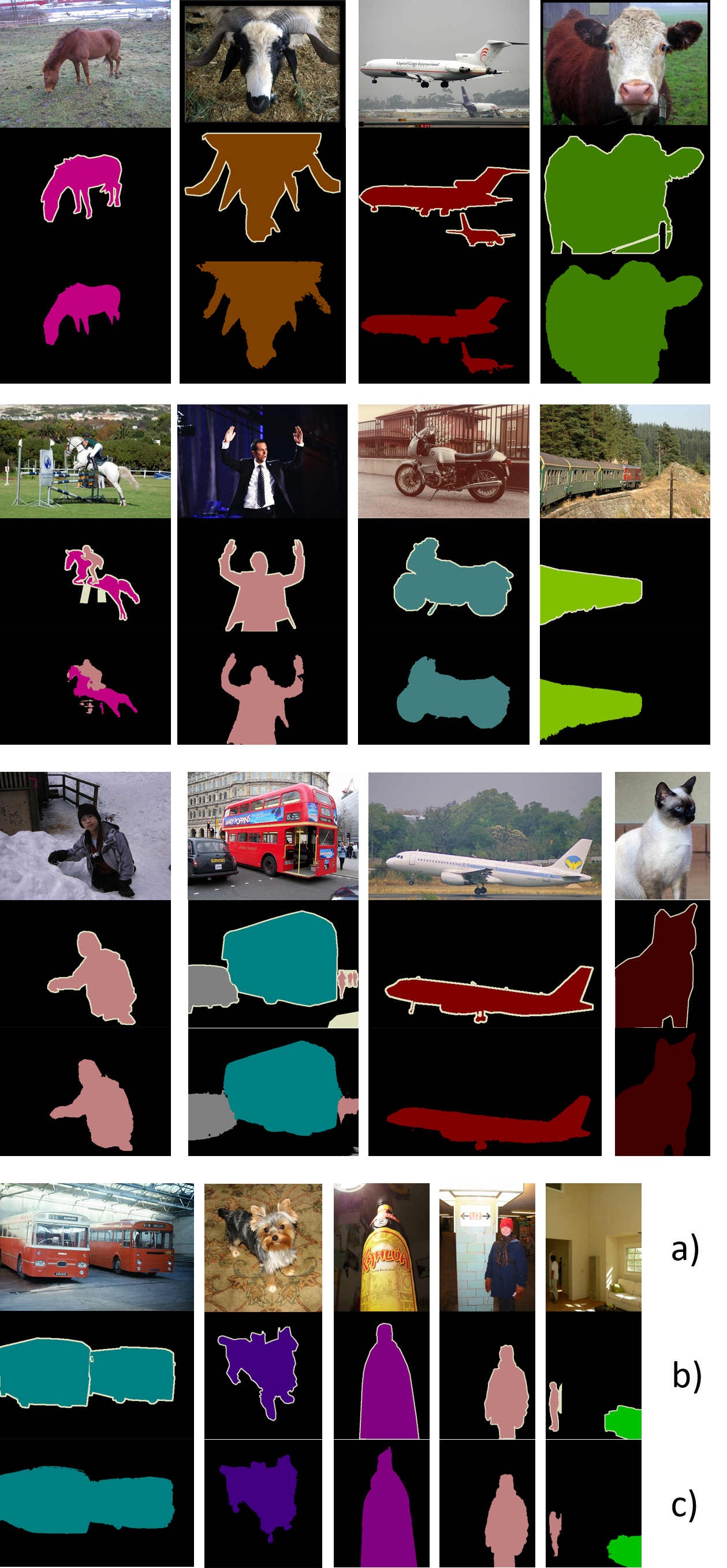}
\caption{Visualization of pseudo-segmentation masks on the PASCAL VOC 2012 \textit{val} set. a) Input image; b) Ground truth; c) CRT}
\label{fig:supp_val}
\end{center}
\end{figure}

\begin{figure*}[t]
\begin{center}
\includegraphics[width=1.0\textwidth]{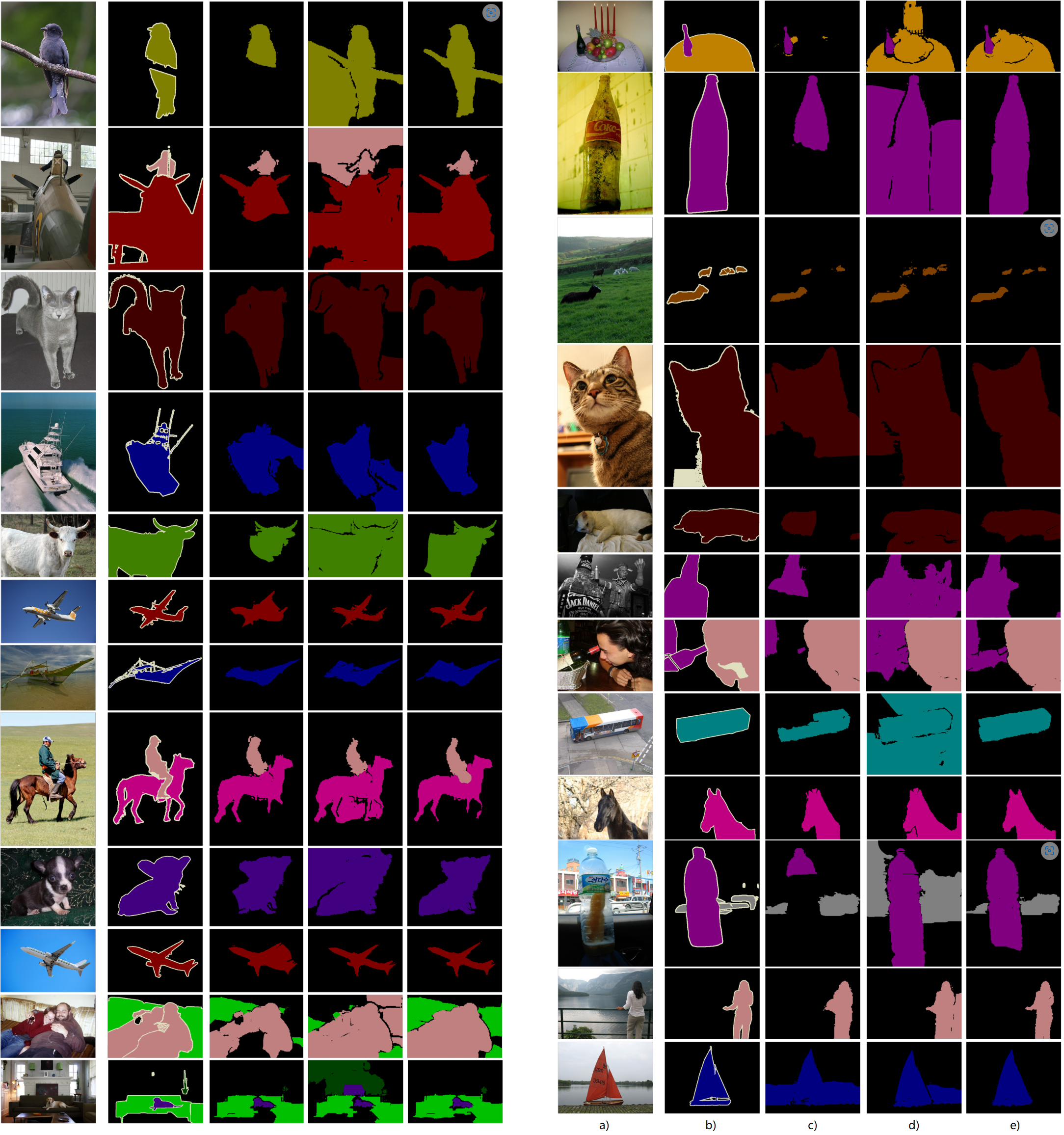}
\caption{Visualization of pseudo-segmentation masks on the PASCAL VOC 2012 \textit{training} set. a) Input image; b) Ground truth; c) IRNet; d) TS-CAM; e) CRT}
\label{fig:supp_mask}
\end{center}
\end{figure*}

To further demonstrate the effectiveness of CRT, we conduct more visualization experiments on the PASCAL VOC 2012 \textit{training} set, as shown in Figure \ref{fig:supp_mask} and the PASCAL VOC 2012 \textit{val} set, as shown in Figure \ref{fig:supp_val}

